\documentclass{article}

\usepackage{arxiv}

\renewcommand{\undertitle}{Published in International Journal for Numerical Methods in Engineering\\126(24) (2025) e70193\\[2pt]
  {\normalfont\small \url{https://doi.org/10.1002/nme.70193}}}
\renewcommand{\headeright}{Published in IJNME 126(24) (2025) e70193}
\usepackage[utf8]{inputenc}
\usepackage[T1]{fontenc}
\usepackage{graphicx}
\usepackage{amsmath,bm}
\usepackage{amssymb}
\usepackage{xcolor}
\usepackage{float}
\usepackage{tikz}
\usepackage{comment}
\usepackage{pgfplots}
\usepackage[font=footnotesize,labelfont=bf]{subcaption}
\usepackage{hyperref}
\usepackage{placeins}
\usepackage{booktabs}
\newenvironment{tablenotes}{\par\footnotesize\begin{list}{}{\setlength{\leftmargin}{0pt}\setlength{\itemindent}{0pt}\setlength{\labelwidth}{0pt}\setlength{\labelsep}{0pt}}}{\end{list}}
\pgfplotsset{compat=1.18}

\graphicspath{{./}}

\def\vv{{\bm{v}}}
\def\ee{{\bm{e}}}

\newcommand{\periodicModelName}{P-GNN}
\newcommand{\proposedModelName}{P-DivGNN}

\begin{document}

\title{Physics-Informed Graph Neural Networks to Reconstruct Local Fields Considering Finite Strain Hyperelasticity}

\author{%
  Manuel Ricardo Guevara Garban$^{1,2}$\thanks{Corresponding author: \texttt{manuel.guevara-garban@u-bordeaux.fr}}\qquad
  Yves Chemisky$^{1}$\qquad
  Michaël Clément$^{2}$\qquad
  Étienne Prulière$^{3}$\\[10pt]
  \mdseries\small
  $^{1}$Univ. Bordeaux, CNRS, Bordeaux INP, I2M, UMR 5295, Talence 33400, France\\
  $^{2}$Univ. Bordeaux, CNRS, Bordeaux INP, LaBRI, UMR 5800, Talence 33400, France\\
  $^{3}$Arts et Metiers Institute of Technology, CNRS, Bordeaux INP, I2M, UMR 5295, Talence 33400, France
}
\date{}

\maketitle

\begin{abstract}
We propose a physics-informed machine learning framework called \textit{\proposedModelName}
to reconstruct local stress fields at the micro-scale, in the context of multi-scale simulation
given a periodic micro-structure mesh and mean, macro-scale, stress values.
This method is based in representing a periodic
micro-structure as a graph, combined with a message passing graph neural
network. We are able to retrieve local stress field distributions, providing average stress values
produced by a mean field reduced order model (ROM) or Finite Element (FE) simulation at the macro-scale. The
prediction of local stress fields are of utmost importance considering fracture analysis or the definition
of local fatigue criteria. Our model incorporates physical
constraints during training to constraint local stress field equilibrium state and
employs a periodic graph representation to enforce periodic boundary
conditions. The benefits of the proposed physics-informed GNN are evaluated
considering linear and non linear hyperelastic responses applied to varying geometries. In the non-linear hyperelastic case, the proposed method achieves significant computational speed-ups compared to FE simulation, making it particularly attractive for large-scale applications.
\end{abstract}


\renewcommand\thefootnote{}
\footnotetext{\textbf{Abbreviations:} GNN, graph neural networks; FEM, finite element method; ROM, reduced order model.}

\renewcommand\thefootnote{\fnsymbol{footnote}}
\setcounter{footnote}{1}

\section{Introduction}\label{sec:intro}

Technological innovation is driving material science to new heights, especially
in the development of modern health devices, vehicles, and renewable energy
systems. These applications demand lightweight, multi-functional components
that maintain mechanical integrity. Additive manufacturing, along with other
emerging processing technologies, has significantly expanded the range of
achievable architectured geometries, potentially overcoming traditional design
constraints. However, designing and optimizing these structures requires
extensive simulations to evaluate mechanical responses under various loading
conditions. Multi-scale optimization that demands multiple iterations, varying
geometry at both the micro and macro-scale as well as material parameters is
particularly costly. Further considering non-linear behaviors such as plasticity and non-linear
kinematics render such optimization process unrealistic with full-field
simulation tools.

\subsection{Multi-scale simulation}
A computationally efficient way to simulate mechanical problems that have two
different characteristic scales (or more) lays in the use of homogenization
theory. If the scales are well separated, we can split the problem into microscopic and
macroscopic parts. These two boundary condition sub-problems
are coupled: Considering appropriate boundary conditions, the microscopic
problem provides the homogenized properties (mean stress, tangent stiffness
matrix) that are required to solve the macroscopic problem. Considering a
linearized two-scale finite element (FE) simulation at each time increment, the
macroscopic problem requires such information at each integration point. The
resolution of the macroscopic problem provides in return the predicted strain
that allows to define the appropriate local boundary condition for the
microscopic one. This method is referred to as the $\mathrm{FE^2}$
method \cite{ref:fe2_feyel1, ref:fe2_feyel2}. Considering linear elastic
response, the microscopic problem, which consists of six linear simulations in 3D
(and three in 2D) for the determination of the homogenized stiffness matrix, need
only to be solved once since the stiffness matrix is time independent. Nevertheless,
if the microstructure varies inside the macroscopic domain (e.g., foam with a
density gradient), a different homogenization problem needs to be solved at
each integration point. Also, considering a design process such as shape
optimization, a high number of microscopic simulations may be required to
determine an optimal shape. In many cases, homogenized properties are not
sufficient in the design process, especially considering the response to strength and fatigue.
In these cases, a local criterion based on the stress components is
required, so the full local stress field needs to be computed.
Note that the computation of full fields comes naturally considering FE$^2$ methods,
which suffers from computational efficiency.
More efficient FE$^2$ have been proposed recently~\cite{ref:lange_monolithic}, however
the computational time is still not compatible with an iterative optimization approach
for multi-scale problems.

Prior to the development of FE$^2$ methods, micro-mechanical approaches have been developed, considering several
approaches~\cite{quFundamentalsMicromechanicsSolids2006} to link the average
local fields of different phases with the global average. Such relation is restricted to specific geometry (i.e.,
ellipsoidal inhomogeneities), while approximations were developed for other geometries, the field
being described as a piecewise field per phase~\cite{Dvorak1992}.

\subsection{Reduced order models}

To mitigate the high costs associated with multi-scale simulations and following micro-mechanical approaches, reduced
order models (ROM) have been proposed to simplify simulations at the
micro-scale level while retaining essential microstructure features and
behaviors, such as mean stress values. Such ROM, that
have added plastic modes to better represent the local fields of non-linear
heterogeneous microstructures, have been developed
in~\cite{michelNonuniformTransformationField2003} and extended
in~\cite{roussetteNonuniformTransformationField2009} considering the use of
Proper Orthogonal Decomposition (POD) to identify those plastic modes. The
plastic modes are therefore utilized to better represent strain localization
that appear in phases subjected to the development of plastic strains.

AI-based approaches were also developed to drastically cut down computational times~\cite{ref:aymen_fe_lstm},
at the cost of retaining only the average stress field and tangent modulus.

Another local field estimation has been introduced in~\cite{LiuBessaLiu.2016}, referred
to has the self-consistent clustering analysis. In this approach, local fields are approximated
as a group of clusters with similar mechanical features. This method is very relevant
to determine the homogenized response of complex heterogeneous structures without
the determination of specific local non-linear deformation modes that are
microstructure-dependent.

Nonetheless, all these methods are not dedicated to the precise definition of local fields, but
rather focus on satisfying, computational efficient, estimation of the global response of non-linear,
multi-scales structures.

\subsection{Deep learning methods}

Recently, Machine Learning (ML) approaches have been extensively employed for surrogate
modeling \cite{ref:ml_stochastic, ref:probabilistic_response,
ref:pinn_surrogate_modeling} and predict full fields, not restricted to the specific case of
multi-scale modeling. Deep Learning methods, such as Convolutional Neural Networks (CNNs) \cite{ref:cnn_legacy}, have been employed to predict full stress fields on 2D geometries, as demonstrated by the framework "StressNet," which utilizes CNNs to estimate stress fields in cantilevered structures \cite{ref:cnn_cantilevered_structures}. Furthermore \cite{ref:gupta_cnn_multiscale_mechanics} proposed to use a U-Net \cite{ref:cnn_unet} architecture as an end-to-end ML-driven framework to accelerate multi-scale mechanics simulations of heterogeneous macro-structures.
The U-Net architecture has also been employed for predicting stress fields in
additively manufactured metals with intricate defect networks
\cite{ref:cnn_stress_fields_additive}.

Other CNN-based methods have explored the use of generative models, such
as Generative Adversarial Networks (GANs) \cite{ref:cnn_gan_legacy} and
Diffusion Models \cite{ref:diffusion_models_legacy}. For example,
\cite{ref:cnn_gan_strain_stress_tensors} proposed a GAN model for predicting
strain and stress tensors in hierarchical composite microstructures. Similarly,
\cite{ref:cnn_stress_estimation_diffusion_model} used Diffusion Models to
estimate stress fields in 2D geometries.

While CNNs have demonstrated success in predicting stress fields in mechanics,
their applicability is often limited when extending to unstructured
geometries. This limitation stems from the need to use a regular grid (2D or 3D)
as input for CNNs, which poses significant challenges when working with meshes
that have varying refinement sizes. In such cases, local features at different
resolutions are essential for accurate predictions, and grid-based representations struggle to capture these multi-scale dependencies. Point-cloud-based methods such as the deep operator network proposed in \cite{ref:point_cloud_deep_onet}, have been introduced to address this limitation by leveraging 1D convolutional layers for field predictions on parameterized geometries. Although, point-cloud-based approaches are inherently limited for field predictions on mesh data due to their inability to explicitly capture topological connectivity and local geometric relationships, both of which are essential for accurately representing multi-scale dependencies and intricate features in stress field predictions.

To overcome the limitations of CNNs in simulating unstructured geometries,
Graph Neural Networks (GNNs) \cite{ref:gnn_legacy} have gained increasing
attention. By operating directly on graph-structured data, such as mesh data,
GNNs can effectively capture spatial relationships and multi-scale dependencies
inherent in unstructured domains, offering a more flexible approach to stress field
prediction. One notable contribution is \textit{MeshGraphNet}, a GNN architecture
proposed by \cite{ref:mesh_graph_net}. This framework adopts an
Encode-Message Passing-Decode architecture, enabling accurate representations of
mesh-based simulations. \textit{MeshGraphNet} has become a state-of-the-art approach for
predicting scalar fields on mesh data \cite{ref:gnn_nature_stress_strain,
ref:gnn_multiscale_periodic, ref:gnn_multiscale_mesh_graph_net}.

Despite these advances, integrating physical constraints such as equilibrium
conditions and periodic boundary conditions into GNN frameworks remains a
challenge. Some exploratory works aim to incorporate physics-informed priors,
also known as Physics-Informed Neural Networks (PINNs) \cite{ref:PINN_legacy},
into GNNs. For instance, \cite{ref:gnn_pde_gnn_embedded} proposed a GNN-based
solver for partial differential equations (PDEs) considering boundary
conditions, while \cite{ref:gnn_thermo_informed} introduced a
thermodynamics-informed GNN to predict the temporal evolution of dissipative
dynamic systems. However, enforcing equilibrium constraints on predicted stress
fields using GNNs remains an unresolved issue.
\cite{ref:divergence_free_nn} Explored the concept of divergence-free neural
networks for fluid mechanics, using automatic differentiation, even so, this
approach remains experimental and difficult to scale on larger meshes due to the expensive use of
automatic differentiation.

\subsection{Contributions}

In this work, we propose a machine learning method as a deep learning based localization operator. As most of the Reduced Order Models for Computational
Homogenization degrades the information about local fields for the sake of
computational efficiency, durability design requires such information to
efficiency be able to predict damage and or local failure. A localization
problem is formulated such that local mechanical field (e.g., stress) can be
determined from the average mechanical stress.

Our contributions are listed as follows: (i) we introduce a novel physics-informed loss function
that constrains the neural network to predict local stress fields that follows
the local equilibrium state condition (i.e. $\mbox{div} \, \mathbf{\sigma} = 0$) during the training phase; (ii) we use a graph periodic representation implemented to enhance the representation of
periodic boundary conditions for the GNN model during training and inference
phases; (iii) validations are performed on a 2D test case
of a hole plate, varying both the radius and position of the hole to further evaluate
the prediction capacity of the proposed model; (iv) the proposed model is tested in reconstructing a local stress field from a
non linear hyperelastic simulation in finite strain showing an important computation speed-up compared to FEM.

\section{Background}

\subsection{Homogenization}

In legacy multiscale approaches, three key components are required: (i) a constitutive model for each phase (ii) the identification of localization tensors which relate the global and local responses and (iii) techniques to transition and bridge the different scales.

Constitutive models will be described in this article considering elastic and non-linear hyperelastic models.
Note that if not fully described to focus on the development of the GNN-based reduced order model, all the models have been developed such that the~\emph{tangent modulus} is defined so that it can be fully integrated in finite element codes to build the ground truth simulation database.

Micromechanical methods utilized in the literature of heterogeneous materials can be classified in two main categories -- \emph{mean-field approaches} and \emph{full-field approaches}.

\emph{Mean-field approaches} is often referred to methods that consider the phase-averaged response of the various constituents. Thoses approaches typically leverage the seminal result of~\cite{Eshelby1957} and his Equivalent Inclusion Method~\cite{Eshelby1957,Mura1987,QuCherk2006} to account for the presence of multiple inhomogeneities.
In a few cases, analytic approaches can be utilized to describe the behavior of RVEs, for example in the case of composites reinforced with spheres or cylinders~\cite{Hashin1962,HashRos1964,ChristLo1979}.

\emph{Full-field approaches} are those in which the position dependent field values are determined and then averaged for the effective, macroscopic response.  The latter category utilize numerical techniques, e.g. the finite element method (FEM), to solve representative computational domains -- repeating unit cells (UC) for periodic materials and representative volume elements (RVEs) for statistically homogeneous materials~\cite{Pinderaetal2009}. Also here, analytical methods exists in the case of periodic laminates, even if the constituents exhibit a non-linear behavior~\cite{ChaChaCheMer2016}.

Note that the identification of the overall response of a composite with periodic microstructure requires most of the times
efficient computational tools like finite element or finite volume methods. To reduce the computational cost of FE calculations, a method that utilizes Fast Fourier Transforms has been
developed \cite{MoulSuq1998}. Analytical or semi-analytical homogenization solutions exist for elastic multilayered composites \cite{KalKol1997} and for elastic fiber composites with square or
hexagonal symmetry \cite{Ramosetal2001,Diazetal2001,WanPin2016,WanPin2016b}.

\subsubsection{Linearized, incremental formulation}\label{sec:linearization}

The relation between the stress, the strain and the temperature is often very complicated in nonlinear materials. For hyperelastic and dissipative mechanisms
where internal variables are involved, considering an incremental formulation adapted to numerical implementation in simulation codes,
such relation is expressed in an implicit way.
The problem has been already linearized in time, using for instance an approach based on the well known return mapping algorithm scheme~\cite{SimoHughes1998}. Details about this approach are provided in~\cite{ChatzigeorgiouEtal2008}.

In multiscale analysis, the two problems (macroscale and unit cell) are solved simultaneously, using an iterative scheme: from the macroscale analysis the macroscopic strains
 are used in the unit cell problem to compute the microscale variables to determine the unknown macroscopic stress. In the following the description of the non-linear, periodic
 homogenization framework closely follows

For the macroscale analysis there are two types of increments for a variable $b$:

\begin{itemize}
 \item The time increment between the steps $n$ and $n+1$,
\begin{equation}
 \Delta \bar{b}^{(n+1)}=\bar{b}^{(n+1)}-\bar{b}^{(n)}.
\end{equation}
 \item The FE iterative increment between the steps $m$ and $m+1$,
\begin{equation}
 \eth\bar{b}^{(n+1)(m)} = \bar{b}^{(n+1)(m+1)}-\bar{b}^{(n+1)(m)} = \Delta\bar{b}^{(n+1)(m+1)}-\Delta\bar{b}^{(n+1)(m)}.
\end{equation}
\end{itemize}

For the microscale analysis there are three types of increments for a variable ${b}$:
\begin{itemize}
 \item The time increment between the steps $n$ and $n+1$,
\begin{equation}
 \Delta{b}^{n+1}={b}^{n+1}-{b}^{n}.
\end{equation}
 \item The macroscopic tangent moduli increment between the steps $m$ and $m+1$,
\begin{equation}
\eth {b}^{(n+1)(m)} = {b}^{(n+1)(m+1)}-{b}^{(n+1)(m)} =
\Delta{b}^{(n+1)(m+1)}-\Delta{b}^{(n+1)(m)}.
\end{equation}
 \item The unit cell iterative increment between the steps $m^*$ and $m^*+1$,
\begin{equation}
 \delta{b}^{(n+1)(m+1)(m^*)} = {b}^{(n+1)(m+1)(m^*+1)}-{b}^{(n+1)(m+1)(m^*)}
= \Delta{b}^{(n+1)(m+1)(m^*+1)}-\Delta{b}^{(n+1)(m+1)(m^*)}.
\end{equation}
\end{itemize}

To avoid lengthy expressions and following~\cite{ChatzigeorgiouEtal2008}, every quantity that refers to the time step
$n+1$ will be presented without index. Only quantities that
refer to the previous time steps are presented with the exponent $(n)$.

\subsubsection{Unit cell problem}

In this section we introduce the following tensor operators using Einstein notation:

\[
\mathbf{C} = \mathbf{A}:\mathbf{B}
\qquad\text{with}\qquad
C_{i_1\,i_2\ldots i_{n+m}}
= \quad
A_{i_1\,i_2\ldots i_n\,pq}\,
B_{pq\,i_{n+1}\,i_{n+2}\ldots i_{n+m}}\,,
\]

\[
\mathbf{C} = \mathbf{A}\,\tilde{:}\,\mathbf{B}
\qquad\text{with}\qquad
C_{i_1\,i_2\ldots i_{n+m}}
= \quad
A_{i_1\,i_2\ldots i_n\,pq}\,
B_{i_{n+1}\,i_{n+2}\ldots i_{n+m}\,pq}\,,
\]

$\mathbf{A}:\mathbf{B}$ is often called the double contraction product. When $\mathbf{A}$ and $\mathbf{B}$ are both second order tensors, the resulting $\mathbf{C}$ is a scalar and the operators $:$ and $\tilde{:}$ are equivalent.

\begin{enumerate}

\item As input, the unit cell problem is informed with the macroscopic strain and macro-strain increment, provided exclusively
by the analysis in the macroscale, which means that they do not iterate during the calculations, and thus there is no
$\delta$ increment in $\bar{\mathbf{\varepsilon}}$.
The microscopic  stress tensor at the time step $n+1$ can be expressed in an iterative linearized fashion as
\begin{equation}
\mathbf{\sigma}^{\textrm{updated}} = \mathbf{\sigma} + \mathbf{D}^{\varepsilon} \colon \delta \mathbf{\varepsilon}, \quad \delta\mathbf{\varepsilon}=\textrm{grad}_{\textrm{sym}}\delta \widetilde{\mathbf{u}},
\end{equation}
where $\mathbf{D}^{\varepsilon}$ denotes the mechanical tangent modulus and $\textrm{grad}_{\textrm{sym}}$ is the symmetric gradient.
To ensure the existence and the uniqueness of solution of the microscale problem,
the fourth order tensor $\mathbf{D}^{\varepsilon}$ need to be positive definite at all times.
Using the above linearized expressions, the micro-equilibrium is written as
\begin{equation}
\textrm{div} \left(\mathbf{\sigma}+\mathbf{D}^{\varepsilon} \colon \textrm{grad} \delta \widetilde{\mathbf{u}}\right)=0,
\label{eq:unit_cell_problems_nonlinear}
\end{equation}
Here the gradient is equivalent to the symmetric gradient since $\mathbf{D}^{\varepsilon}$ is assumed symmetric. These conservation laws are accompanied by appropriate microscopic constitutive laws to update the current stress $\mathbf{\sigma}$. With regard to boundary conditions,
periodicity of $\delta \widetilde{\mathbf{u}}$, and antiperiodicity of $\mathbf{\sigma} \cdot \mathbf{n}$ are imposed. The global system of equations
is solved iteratively using a Newton-Raphson Method.
When the convergence is achieved (i.e. when $\delta \widetilde{\mathbf{u}}$ is almost zero), the time increments and the actual values of the
fluctuating term $\widetilde{\mathbf{u}}$ are evaluated, allowing all the microscopic quantities and their averaged (macroscopic) counterparts
to be computed. Moreover, the tangent modulus $\mathbf{D}^{\varepsilon}$ at all microscopic points can be calculated when the convergence is achieved, using
the local constitutive models. These moduli are essential to obtain their macroscopic counterparts in the second part of the unit cell problem.
\item Once the first part is completed, the residual terms of the microscopic stress components are assumed exactly zero.
The strain and the equilibrium are written in the linearized, incremental form as
\begin{equation}
\eth \mathbf{\varepsilon}=\eth\bar{\mathbf{\varepsilon}}+\textrm{grad}_{\textrm{sym}}\eth \widetilde{\mathbf{u}}, \quad  \textrm{div} \eth \mathbf{\sigma}=\mathbf{0},
\label{eq:linearized_microproblem}
\end{equation}
With regard to boundary conditions, periodicity of $\eth \widetilde{\mathbf{u}}$ and antiperiodicity of $\eth \mathbf{\sigma} \cdot \mathbf{n}$ are imposed.
Considering the mechanical tangent moduli, the increment
of the stress tensor is expressed:
\begin{equation}
\eth \mathbf{\sigma}=\mathbf{D}^{\varepsilon}\colon\eth \mathbf{\varepsilon}
\label{eq:linearized_microconstitutive_laws}
\end{equation}

Combining equations \eqref{eq:linearized_microproblem},  \eqref{eq:linearized_microconstitutive_laws} yields
\begin{equation}
\textrm{div} \left(\mathbf{D}^{\varepsilon} \colon \eth \bar{\mathbf{\varepsilon}}
+ \mathbf{D}^{\varepsilon} \colon \textrm{grad} \eth \widetilde{\mathbf{u}} \right) = \mathbf{0},
\label{eq:tangent_moduli_problems_nonlinear}
\end{equation}
The solution of the above homogeneous system (up to an arbitrary macroscopic constant)
is written in the form \cite{Ene1983}
\begin{equation}
\eth \widetilde{\mathbf{u}} = \eth \bar{\mathbf{\varepsilon}} \colon \mathbf{\chi}^{\varepsilon}
\label{eq:microfluctuations}
\end{equation}
where the third order tensor $\mathbf{\chi}^{\varepsilon}$ is periodic and is referred to as corrector term. Substituting \eqref{eq:microfluctuations} in
\eqref{eq:tangent_moduli_problems_nonlinear} yields
\begin{equation}
\eth\bar{\mathbf{\varepsilon}}\colon \textrm{div} \left(\left[\mathbf{D}^{\varepsilon}
+\mathbf{D}^{\varepsilon} \tilde{:} \textrm{grad} \mathbf{\chi}^{\varepsilon}\right]^T\right) = \mathbf{0},
\end{equation}
The last expressions are valid for arbitrary values of the macroscale variables $\eth\bar{\mathbf{\varepsilon}}$,
only if the corrector terms satisfy the linear equations:
\begin{equation}
\textrm{div}\left(\left[\mathbf{D}^{\varepsilon} + \mathbf{D}^{\varepsilon} \tilde{:} \textrm{grad} \mathbf{\chi}^{\varepsilon}\right]^T\right)=\mathbf{0}
\label{eq:unit_cell_problems_tangent}
\end{equation}
Using the solution \eqref{eq:microfluctuations}, the increments of the other microscopic
fields are written as
\begin{equation}
\begin{split}
\eth\mathbf{\varepsilon} & = \mathbf{A}^{\varepsilon}\colon\eth \bar{\mathbf{\varepsilon}} \\
\eth\mathbf{\sigma} & = \mathbf{D}^{\varepsilon}\colon \mathbf{A}^{\varepsilon} \colon\eth \bar{\mathbf{\varepsilon}}
\label{eq:microscopic_fields}
\end{split}
\end{equation}
where
\begin{equation}
\mathbf{A}^{\varepsilon} = \mathbf{\mathcal{I}} + \mathbf{\mathcal{I}} \tilde{:} \textrm{grad} \mathbf{\chi}^{\varepsilon},
\label{eq:macroscopic_tangent_concentrations}
\end{equation}
is the strain localization tensor~\index{localization tensor}.
It is recalled that $\mathbf{\mathcal{I}}$ denotes the fourth order symmetric identity tensor and $\mathbf{I}$ is the second order identity tensor. Averaging of these fields over the unit cell yields
the macroscopic tensors
\begin{equation}
\bar{\mathbf{D}}^{\varepsilon} = \left\langle \mathbf{D}^{\varepsilon} \colon \mathbf{A}^{\varepsilon} \right\rangle
\label{eq:macroscopic_tangent}
\end{equation}
\end{enumerate}
This yields the definition of a localization operator $\mathbf{A}^{\varepsilon}$ that allows to retrieve the local microscopic deformation increment from the knowledge of the
average deformation tensor increment. While in practice, during finite element-based simulation of the non-linear response of a unit cell, providing periodic boundary conditions,
such localization tensor is never evaluated, all methods that actually provide the tangent modulus $\bar{\mathbf{D}}^{\varepsilon}$ are able to provide those localization
tensors at every gauss point of the considered periodic mesh. A localization tensor field can therefore be constructed at node points, relying on the same interpolation functions
and averaging methods that those utilized for the evaluation of the stress field.

The stress localization operator $\mathbf{B}^{\varepsilon}$ is deduced from $\mathbf{A}^{\varepsilon}$ as:

\begin{equation}
\begin{split}
\eth\mathbf{\sigma} & =  \mathbf{D}^{\varepsilon} \mathbf{A}^{\varepsilon} \colon \eth \bar{\mathbf{\varepsilon}} \\
\eth\mathbf{\sigma} & =  \mathbf{D}^{\varepsilon} \mathbf{A}^{\varepsilon} \left( \bar{\mathbf{D}}^{\varepsilon} \right)^{-1} \colon \eth \bar{\mathbf{\sigma}} \\
\eth\mathbf{\sigma} & =  \mathbf{B}^{\varepsilon} \colon \eth \bar{\mathbf{\sigma}} \,, \quad \mathbf{B}^{\varepsilon} = \mathbf{D}^{\varepsilon} \mathbf{A}^{\varepsilon} \left( \bar{\mathbf{D}}^{\varepsilon} \right)^{-1}
\end{split}
\end{equation}

The proposed method in this paper can be viewed, from this perspective, as an AI-based model of this localization operator, considering that the local stress field are directly evaluated at the graph (or mesh) node points from the information of the average stress tensor.

On the other hand, the aforementioned method is implemented in the finite element code \textit{Fedoo}, such that non-linear constitutive laws can be utilized in a non-linear incremental periodic boundary condition framework.

\subsection{Mechanical problem}\label{sec:mechanical_problem}

The mechanical problem addressed here is the numerical evaluation of the response of
multi-scale structures, considering a periodic arrangement of a representative pattern. Assuming a bridging of scale, i.e. that the local cell characteristic size is several decades smaller than the structure, the periodic homogenization theory is adapted to determine an effective response of the representative unit cell (RUC). Within such a unit cell, the local static stress equilibrium equation in the absence of volume forces writes:
\begin{equation}\label{eq:div_sigma_eq_0}
    \mbox{div} \, \sigma = 0 ,
\end{equation}

which reduced in 2D:

\begin{equation}\label{eq:div_sigma_continu}
\mbox{div} \, \sigma
    =  \begin{pmatrix} \frac{\partial \sigma_{xx}}{\partial x} + \frac{\partial \sigma_{xy}}{\partial y} \\ \frac{\partial \sigma_{xy}}{\partial x} + \frac{\partial \sigma_{yy}}{\partial y}\end{pmatrix} = 0,
\end{equation}

where $\sigma$ is the stress tensor and $\mbox{div}$ is the divergence operator.

To solve this problem using a displacement-based approach we need:
(i) a constitutive equation to define the material's response (ii) a kinematical relation that defines the strain field from the displacement field $\mathbf{u}$; (iii) appropriate prescribed boundary conditions.

In the following, the constitutive law is considered either hyperelastic or linear elastic. In this last case, small strain approximation is utilized and the relationship between $\sigma$ and the small strain tensor $\varepsilon$ may be written as:

\begin{equation}\label{eq:constitutive_law}
    \sigma = \mathbf{L}\, : \, \varepsilon ,
\end{equation}
where $\mathbf{L}$ is fourth order the linear elastic stiffness tensor.

Further considering such linearized kinematics, the relation between $\varepsilon$ and $\mathbf{u}$ is:
\begin{equation}\label{eq:strain_def}
    \varepsilon = \frac{1}{2} \left( \nabla \mathbf{u} + \nabla^T \mathbf{u} \right),
\end{equation}

where $\nabla \mathbf{u}$ denotes the displacement gradient.
As for the boundary conditions, considering that we have a periodic microscopic domain, we naturally use the  periodic boundary conditions \cite{ref:suquet_periodic}, that can be written by:
\begin{equation}
\mathbf{u}(\mathbf{x}) =\bar{\varepsilon} \cdot \mathbf{x} + \mathbf{\tilde{u}}(\mathbf{x}) ,
\end{equation}
where $\mathbf{x}=\begin{pmatrix} x\\ y \end{pmatrix}$ is the coordinates vector in the unit cell, assumed in 2D for sake of readability without loss of generality, that is described by the geometric domain $\Omega$.  $\bar{\varepsilon} = \frac{1}{V} \int_{\Omega} \varepsilon \, dV$ is the mean strain tensor over the representative unit cell and $\mathbf{\tilde{u}}$ is a periodic fluctuation, i.e. for two points that belong to opposite faces whose coordinates are noted respectively $\mathbf{x}^+$  and $\mathbf{x}^-$ we have $\mathbf{\tilde{u}}(\mathbf{x}^+) = \mathbf{\tilde{u}}(\mathbf{x}^-)$ and therefore it comes the condition:
\begin{equation}
    \mathbf{u}(\mathbf{x}^+) - \mathbf{u}(\mathbf{x}^-) =  \bar{\varepsilon} (\mathbf{x}^+ - \mathbf{x}^-) ,
\end{equation}
that we have to enforce over the boundary of the domain for each pair of opposite faces. Note that for the microscopic strain localization problem, the mean field $\bar{\varepsilon}$ is supposed to be known considering that the macroscopic structure mechanical problem has been solved.
Note that practical details about the implementation of this kind of conditions inside a FE code can be found in \cite{ref:chemisky_thermo_fem}.

To solve Equations \eqref{eq:div_sigma_eq_0}, \eqref{eq:constitutive_law} and \eqref{eq:strain_def} using the FE method we need to use the corresponding weak formulation that is:
\begin{equation}\label{eq:weak_form}
    \int_\Omega \sigma(\mathbf{u}) : \varepsilon(\mathbf{u^\star}) \, d\Omega = 0 ,
\end{equation}

$\mathbf{u^\star}$ being a kinematically admissible displacement field. We assume a FE discretization with $n$ nodes that allow to build an arbitrary scalar field $f$ from
its nodal values such as:
\begin{equation}\label{eq:fe_discretization}
f(\mathbf{x})
    = \begin{pmatrix} N_1(\mathbf{x}) & N_2(\mathbf{x}) & \cdots & N_n(\mathbf{x}) \end{pmatrix} . \begin{pmatrix} f_1 \\ f_2 \\ \vdots \\ f_n \end{pmatrix} .
\end{equation}
$N_i(\mathbf{x})$ are the classical FE shape functions associated to node $i$ and $f_i$ are the nodal values of $f$. This discretization is used to approximate the displacement field in the context of Equation \eqref{eq:weak_form}.

In the more general framework of finite strains, considering a nearly incompressible hyperelastic material, a strain energy density is defined as the sum of a volume-changing part and a volume-preserving part:

\begin{equation}
\Psi = U(J) + W(\bar{I}_1,\bar{I_2})\,.
\end{equation}

In the above, $J = \textrm{Det} \mathbf{F}$ is the determinant of the deformation gradient $\mathbf{F}$, while $\bar{I}_1,\bar{I_2}$ are the first and second invariants of the volume-preserving right Cauchy-Green strain tensor $\bar{\mathbf{C}}$, that derives from the right Cauchy-Green strain tensor $\mathbf{C}$:

\begin{equation}
\begin{split}
\bar{I}_1 & = J^{-2/3} I_1\,\quad I_1 = \mathrm{tr} (C) \\
\bar{I}_2 & = J^{-4/3} I_2\,\quad I_2 = \frac{1}{2} \left(\mathrm{tr} (C) \mathrm{tr} (C) - \mathrm{tr} (C^2) \right).
\end{split}
\end{equation}

Following \cite{ref:hyperelastic}, the Cauchy stress is written as:

\begin{equation}\label{eq:sigmahyper}
\begin{split}
& \mathbf{\sigma} = p \mathbf{1} + \mathbf{s} \\
& p = \frac{\partial U}{\partial J} \\
& J \mathbf{s} = 2 \frac{\partial W}{\partial \bar{I}_1} \mathrm{dev}\,\bar{\mathbf{b}} + 2 \frac{\partial W}{\partial \bar{I}_2} \left( \mathrm{tr}\,\bar{\mathbf{b}}\,\mathrm{dev}\,\bar{\mathbf{b}} - \mathrm{dev}\,\bar{\mathbf{b}}^2 \right).
\end{split}
\end{equation}

In Equation~\eqref{eq:sigmahyper}, $\bar{\mathbf{b}} = J^{-2/3} \mathbf{b}$ is the volume preserving part of the left Cauchy-Green deformation tensor $\mathbf{b} = \mathbf{F} \mathbf{F}^T$. Also, $p$ denotes the hydrostatic pressure and $\mathbf{s}$ is the deviatoric part of the Cauchy stress $\mathbf{\sigma}$.
In the following, the nearly incompressible Neo-Hookean hyperelastic law is utilized, with:

\begin{equation}\label{eq:hyper_elastic_law}
\Psi = \kappa \left( J \, \textrm{ln} J - J + 1 \right) + \frac{\mu}{2} \left(\bar{I}_1 - 3 \right)\,,
\end{equation}

In finite strain, periodic boundary conditions arise naturally from the discretization of the displacement gradient over the unit cell, i.e:

\begin{equation}\label{eq:finite_strain_bc}
    \mathbf{u}(\mathbf{x}^+) - \mathbf{u}(\mathbf{x}^-) = \overline{\nabla u} \, \left(\mathbf{x}^+ - \mathbf{x}^- \right).
\end{equation}

In this case, the FE method requires to solve a non-linear system due to the non-linear relation between strain quantities and the displacement gradient. Hence, a Newton-Raphson algorithm is utilized and an incremental, linearized weak formulation is defined. The geometric nonlinearities are considered based on an updated Lagrangian approach.
The constitutive model utilized in this work is implemented in the open-source library \textit{Simcoon} and the non linear solver is implemented in the finite element library \textit{Fedoo}, all being part of the 3MAH set~\cite{ref:3mah}.

Considering the FE method, the local equilibrium equation (Equation~\eqref{eq:div_sigma_eq_0}) is enforced in a weak sense with respect to the chosen approximation of the displacement field. The stress divergence will therefore not be exactly 0, and we can even use the  stress divergence values as a FE approximation error indicator.

\subsection{Discrete divergence operator} \label{sec:divergence}

Computing the stress divergence values can be a complex task since the stress is often poorly approximated inside an element. For instance, for 2D linear triangle elements, the approximated stress, which is linked to the displacement gradient considering a linear elastic response, is constant over the elements, and therefore the stress divergence is always zero-evaluated. Another way to better approximate the stress divergence is to get the stress values at nodes and differentiate the FE interpolation function to compute a divergence operator from the nodal values. The construction of a stress divergence operator is briefly summarized in the following.

Once the solution is obtained, i.e. the displacement field is known (or displacement increment for incremental problems), it is easy to compute the strain and stress tensors fields from the FE discretization (Equation \eqref{eq:fe_discretization}) at any points in the domain using the strain/displacement relation and the constitutive law (i.e. Equations \eqref{eq:strain_def} and \eqref{eq:constitutive_law}  in our case). The strain and stress fields are generally not continuous between elements, and the nodal values for stress and strain are typically determined by averaging the values from the adjacent elements.

We define the divergence operator as a matrix \(\mathbf{D}\) that computes the divergence at any point in the domain from the nodal values of a 2D vector field using the FE approximation.
\(\mathbf{D}\) is constructed to account for the contributions of the derivatives of the shape functions at each node. That gives:
\[
\mathbf{D}(\mathbf{x}) = \begin{pmatrix}
 \mathbf{D}_x(\mathbf{x})  &  \mathbf{D}_y(\mathbf{x})
\end{pmatrix},
\]
where \(\mathbf{D}_x\) and \(\mathbf{D}_y\) are matrices containing the derivatives of the shape functions with respect to \(x\) and \(y\), respectively:
\begin{equation}
\mathbf{D}_x(\mathbf{x}) =
\begin{pmatrix} \frac{\partial N_1(\mathbf{x} )}{\partial x} & \frac{\partial N_2(\mathbf{x} )}{\partial x} & \cdots &  \frac{\partial N_n(\mathbf{x} )}{\partial x} \end{pmatrix}
\end{equation}
and
\begin{equation}
\mathbf{D}_y(\mathbf{x} ) =
\begin{pmatrix} \frac{\partial N_1(\mathbf{x})}{\partial y} & \frac{\partial N_2(\mathbf{x})}{\partial y} & \cdots &  \frac{\partial N_n(\mathbf{x} )}{\partial y}\end{pmatrix}.
\end{equation}

In the same way, a divergence operator $\mathbf{D_{\sigma}}$ is built to compute the approximated stress divergence from nodal values:
\[
\mathbf{div}(\sigma(\mathbf{x})) = \mathbf{D_{\sigma}}(\mathbf{x}) \cdot \boldsymbol{\sigma},
\]
where \(\boldsymbol{\sigma}\) is the vector of nodal stress components, that is built by stacking the nodal values of each stress component such as:

\begin{equation}
\boldsymbol{\sigma} = \begin{pmatrix}
[\sigma_{xx} ] \\ [ \sigma_{yy} ] \\ [ \sigma_{xy} ]
\end{pmatrix}
\end{equation} and

\begin{equation}
\mathbf{D}_{\sigma}(\mathbf{x}) = \begin{pmatrix}
\mathbf{D}_x(\mathbf{x}) & 0 &  \mathbf{D}_y(\mathbf{x}) \\
0 & \mathbf{D}_y(\mathbf{x})  &  \mathbf{D}_x(\mathbf{x})
\end{pmatrix}.
\end{equation}

Like the stress and strain fields, the divergence computed with this operator is not continuous between elements. The nodal divergence field is calculated by averaging the values at adjacent elements.
In practice we build a discrete divergence matrix operator  $\hat{\mathbf{D}}_{\sigma}$ of size $(2n,3n)$, that directly computes the divergence nodal values ($2n$ values since there are two divergence terms per node) from the nodal stress values (3 components of stress on each node).

The divergence of the stress field at node $i$ can be expressed as:
\begin{equation}\label{eq:div_operator}
\left. \mathbf{div} (\boldsymbol{\sigma}) \right|_{\mathbf{x} = \mathbf{x}_i} = \begin{pmatrix}(\hat{\mathbf{D}}_{\sigma} \cdot \boldsymbol{\sigma})_i & (\hat{\mathbf{D}}_{\sigma} \cdot \boldsymbol{\sigma})_{n+i} \end{pmatrix},
\end{equation}

where $ \left. \mathbf{div} (\boldsymbol{\sigma}) \right|_{\mathbf{x} = \mathbf{x}_i} \in \mathbb{R}^2$ is a 2D vector containing the two terms of the divergence, presented in Equation \eqref{eq:div_sigma_continu} for a given node $i$ at coordinates $\mathbf{x}_i$. In this last expression, the index $n+i$ is related to the second term of the divergence vector at node $i$.

This discrete operator provides a practical method to evaluate the stress divergence across the computational domain and enables its use in error estimation or for iterative improvement of the FE solution. The construction of \(\mathbf{D}\) depends on the chosen element type and the corresponding shape functions.

In this work, we propose to use this discrete operator to compute a physics-informed loss that constrains the predictions of the deep learning framework during the training phase to adhere to the equilibrium state condition.

\subsection{Graph Neural Networks}\label{sec:gnn_mp}

The deep learning model used for this study is referred as Graph Neural Networks
(GNN), this model operating on graph data as input. Graphs are an abstract
representation of relational data defined as a set of vertices (or nodes) noted $\mathcal{V} =
\{v_1, v_2, ..., v_n\}$ and a set of unordered pairs of vertices noted $\mathcal{E}$ called
edges representing connections or relationships between vertices.  Each edge $e
\in \mathcal{E}$ is an element of the form $(v_i,v_j)$. We can associate a feature node
vector $\vv_i \in \mathbb{R}^d$ of dimension $d$ to each node $v_i \in \mathcal{V  }$ and a
feature edge vector $\ee_{ij} \in \mathbb{R}^k$ to each edge $(v_i,v_j) \in \mathcal{E}$ of dimension $k$. The
feature vectors can be used to encode various properties of nodes and edges, such as
categorical information or numerical measurements.

GNNs rely on a mechanism called Message Passing Layers
(MPL), where each message passing operation consists in propagating and aggregating
information across nodes in a graph. The goal of this mechanism is to
iteratively update the features of each node by exchanging information "messages" with its neighbors. Information that is stored in edges can participate to the message passing operation.

A message passing layer is a local operation that is performed on a given node $v_i$
taking into account the information of its set of neighbors $j \in
\mathcal{N}(i)$ and the corresponding edge information $e_{ij}$.

The Message Passing GNN can be seen as a generalization of Convolutional Neural
Networks \cite{ref:cnn_legacy}, since both methods operate locally. However, GNNs are
not restricted to grid data as is the case of CNN. GNN can indeed be applied to
very general structures of data, which can be represented as graphs. Finite
Element meshes are particularly adapted to a graph representation, since they
share structural characteristics as they both involve nodes (vertices) and
connections (edges). The latter represents the geometric connectivity between
nodes. GNNs appear as a very adapted framework for the task of local stress
field reconstruction. Moreover, FE mesh data can present different
resolutions in terms of local mesh refinement, which is difficult to represent
in a grid based approach.

\section{Local Equilibrium Constraint Graph Neural Networks for local stress reconstruction}\label{sec:method}

\begin{figure*}[htpb]
    \centering
    \includegraphics[width=0.9\textwidth]{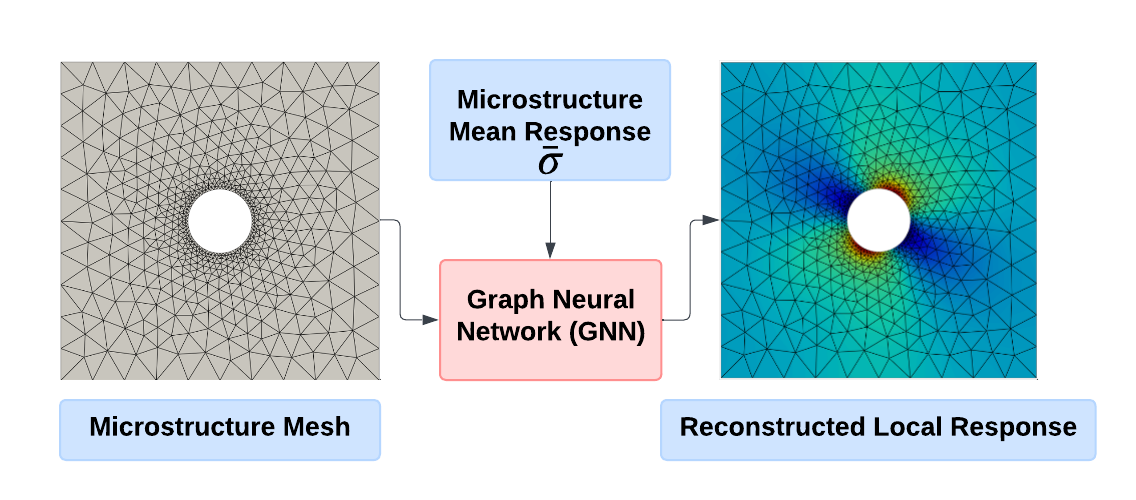}
    \caption{Overview of the proposed local stress reconstruction method. The mean microstructure stress response $\bar{\sigma}$, is uniformly assigned to all nodes of the input mesh. A Graph Neural Network (GNN) then acts as a localization operator, predicting the corresponding local stress field $\sigma_i$ at each node.}
    \label{fig:schema_lsr}
\end{figure*}

The Graph Neural Network will be applied next as a localization model, with the task of reconstructing the local stress field $\sigma$ on a periodic mesh $\mathcal{M}$, given only the mean mechanical response $\bar{\sigma}$ (see Figure \ref{fig:schema_lsr}). Local field reconstruction is understood here as predicting the stress field values at each node of the mesh, i.e. at each vertex of the corresponding GNN. Note that since this graph corresponds to a FE mesh, interpolation functions can be used to reconstruct a continuous piecewise mechanical field.

The microstructure mesh is therefore represented as a graph that encodes topological features such as node positions and mean mechanical properties. The mean mechanical response $\bar{\sigma}$ is assigned as a uniform attribute to all nodes, while topological features, such as node positions, vary across nodes. Edge information is represented by the Euclidean distance between nodes. To account for periodicity, we introduce an additional preprocessing step in which edges are added between boundary nodes. This approach significantly enhances message passing within the model, giving improved results that will be presented in Section \ref{sec:results_linear}. The model incorporating periodic edges is referred to as \textit{\periodicModelName}, which serves as the baseline model in place of a classical GNN.

Node and edge features of the input graph are projected to a latent representation using a Multi-Layer Perceptron (MLP) encoder. The resulting latent graph is processed by a Message Passing GNN framework (as described in Section \ref{sec:gnn_mp}), and finally, the latent graph is decoded by an MLP decoder, producing local stress values as output. This GNN architecture, commonly known as \textit{MeshGraphNet}, has been successfully applied to scalar fields predictions on mesh-based data\cite{ref:gnn_nature_stress_strain, ref:mesh_graph_net, ref:gnn_multiscale_mesh_graph_net} and represents the state-of-the-art in this domain.

As most methods rely on the comparison with numerical simulation as ground truth, we recall here that the mechanical equilibrium is satisfied in a weak sense, i.e. considering the volume integral of the domain. Additionally, we propose a physics-informed loss function that constrains the network to predict local stress fields that satisfy the local static stress equilibrium condition described in Section \ref{sec:mechanical_problem}. The model that implements this physics-informed loss is referred to as \textit{\proposedModelName}.

Section \ref{sec:mesh_graph_repr} provides a detailed description of how the microstructure mesh is represented as a graph. Section \ref{sec:implemented_model} explains the Encode-Message Passing-Decode mechanism, and Section \ref{sec:divergence_loss} presents the proposed physics-informed loss used during training.
\subsection{Representing mesh as a graph}\label{sec:mesh_graph_repr}

    \begin{figure}[ht]
        \centering
        \includegraphics[width=0.6\textwidth]{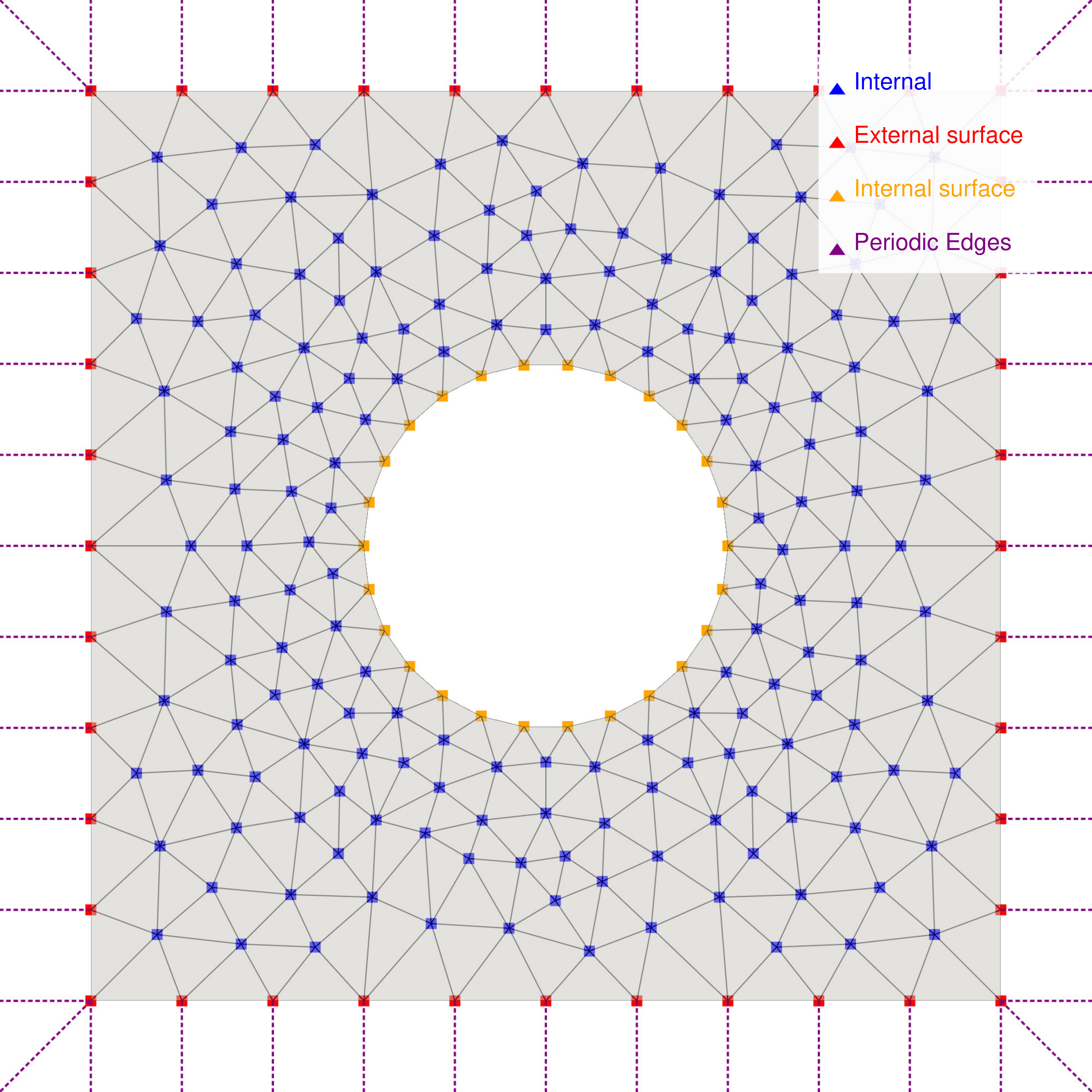}
        \caption{Illustration of the graph representation of a mesh, node labeling in function of topological features and periodic edges added to reinforce message passing.}
        \label{fig:graph_repr_schema}
    \end{figure} \FloatBarrier

To construct a graph representation of the periodic mesh, mesh nodes are labeled according to their position. As shown in Figure \ref{fig:graph_repr_schema}, nodes are classified into internal nodes, internal surface nodes, and external surface nodes using a label $\alpha$. To enhance message passing, additional edges are introduced between boundary nodes. The feature information for these additional edges is set to zero. An alternative method for representing periodicity is to merge boundary nodes, but this approach makes it challenging to assign Cartesian positions to the affected nodes.

We define $\mathcal{M} = (\mathcal{V}, \mathcal{E})$ as an undirected homogeneous graph representing a FE mesh. Each node $v_i \in \mathcal{V}$ is associated with a feature vector $\vv_i = (\bar{\sigma}, \mathbf{x}_i, \alpha_i)$, where:

$\bar{\sigma} \in \mathbb{R}^3$ represents the mean stress values $(\bar{\sigma}_{xx}, \bar{\sigma}_{yy}, \bar{\sigma}_{xy} )$ uniformly assigned to all nodes.

$\mathbf{x}_i \in \mathbb{R}^2$ denotes the spatial coordinates of node $v_i$.

$\alpha_i \in \{ -1, 0, 1 \}$ is an identifier indicating whether the node belongs to an internal surface, an internal node, or is subject to periodic boundary conditions (as illustrated in the legend of Figure \ref{fig:graph_repr_schema}).

Each edge $(v_i, v_j) \in \mathcal{E}$ is associated with a feature vector $\ee_{ij} \in \mathbb{R}$ that encodes the Euclidean distance $\textrm{dist}(\mathbf{x}_j, \mathbf{x}_i) \in \mathbb{R}$ between the spatial coordinates of nodes $v_i$ and $v_j$. To account for periodicity, additional edges are introduced between boundary nodes, with the edge feature vector for these edges set to zero. This periodic edge enhancement is visualized in Figure~\ref{fig:graph_repr_schema}, where purple edges represent the added periodic edges. This approach facilitates more effective message passing and improves the model's ability to capture periodicity within the mesh.

\subsection{Implemented Model}\label{sec:implemented_model}

\begin{figure}[ht]
    \centering
    \includegraphics[width=\textwidth]{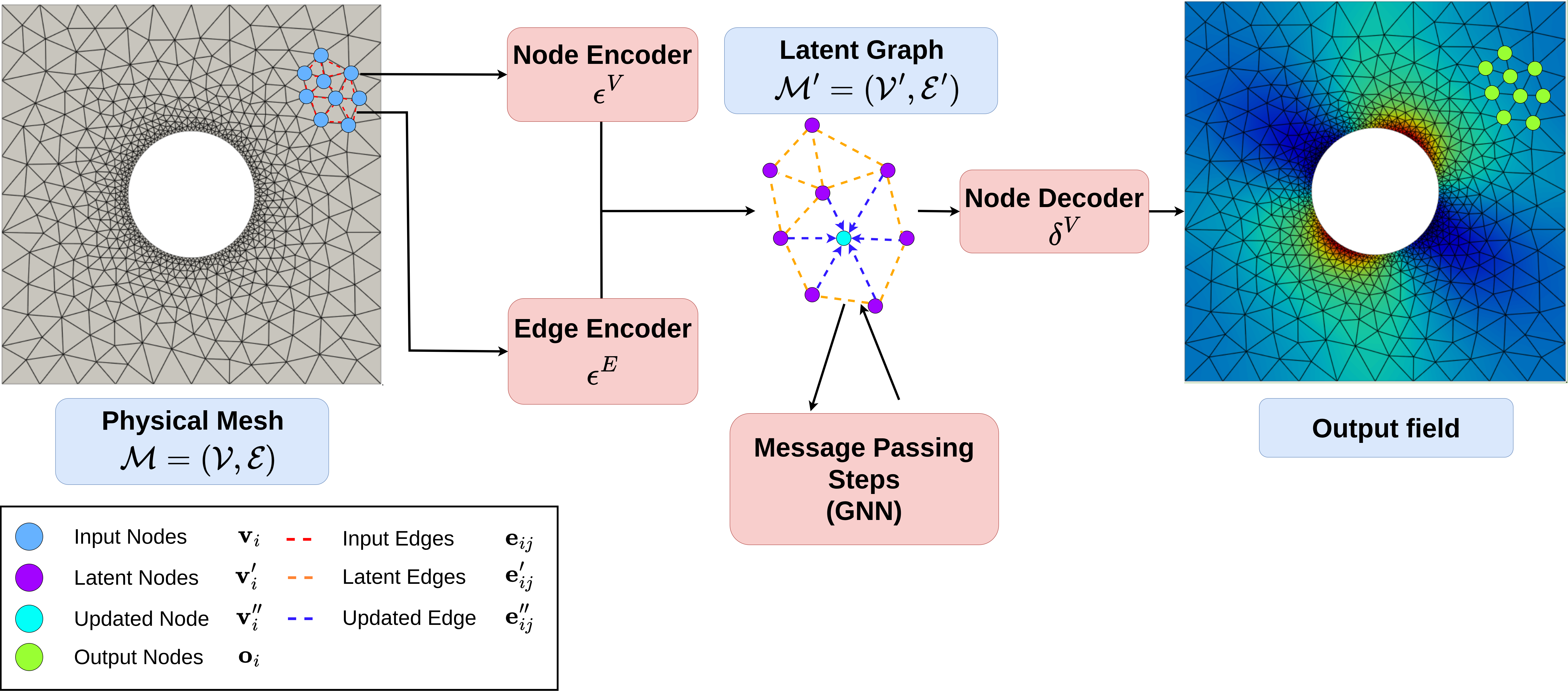}
    \caption{Schematics of the proposed model illustrating \textit{Encode-Message Passing-Decode} architecture. Nodes $\vv_i$ and edges $\ee_{ij}$ of the input graph are embedded through the $\epsilon^V$ and $\epsilon^E$ encoders respectively. This embedded graph is then processed by a GNN which will update the nodes and edges embeddings through message-passing steps. Finally, the node embeddings are decoded with  $\delta^V$ to obtain the output field. Here, only a small subgraph is shown for visualisation purposes, but in practice the whole graph is processed by the architecture.}
    \label{fig:message_passing_schema}
\end{figure}

As illustrated in Figure \ref{fig:message_passing_schema} the model implemented for this study follows a \textit{Encode-Message Passing-Decode} architecture, a framework that has been successfully used to predict stress, strain, and displacement fields in mesh data \cite{ref:mesh_graph_net, ref:gnn_nature_stress_strain, ref:gnn_multiscale_periodic}.

 \FloatBarrier

The proposed GNN model is used as a mapping function:

\begin{equation}
\textrm{GNN}(\vv_i, \mathcal{M}) \mapsto (\hat{\sigma}_{xx_i}, \hat{\sigma}_{yy_i}, \hat{\sigma}_{xy_i})
\end{equation}

allowing to estimate local stress fields values $\hat{\sigma_i}$ for each node $\vv_i$ using mesh structure features represented as a graph $\mathcal{M}$.

The architecture of the GNN consists of three components:

\textbf{Encoder}: The encoder processes the input node feature vectors $\vv_i$, and edge feature vectors $\ee_{ij}$, projecting them into corresponding latent spaces,  $\vv_i' \in \mathbb{R}^H$ and \(\ee_{ij}' \in \mathbb{R}^H\) with \(H\) the dimension of the latent space. This transformation is achieved using two independent multi-layer perceptrons (MLPs): \(\epsilon^{V}\) for node features and \(\epsilon^{E}\) for edge features. The transformations are defined as follows:

\begin{equation}
    \vv_i' = \epsilon^{V}(\vv_i), \quad  \ee_{ij}' = \epsilon^{E}(\ee_{ij}),
\end{equation}

Using the transformed features, the embeddings of the input graph $\mathcal{M}$ are updated giving a latent graph noted \(\mathcal{M}' = (\mathcal{V}', \mathcal{E}')\). It is important to emphasize that this encoding process operates solely on the feature vectors of the graph and does not consider its structural information and the topology of the graph is not changed neither.

\textbf{Message Passing}: This module operates on the latent graph \(\mathcal{M}'\), incorporating the graph's structural features (e.g., the connections between nodes) through \(L\) message-passing steps. During each step, the module updates the edge and node feature vectors using two separate multi-layer perceptrons: \(\phi\) for edge features and \(\gamma\) for node features. For a given message step layer $l \in [1, 2, \dots, L]$ the transformations are defined as follows:

The edge feature vectors \(\ee_{ij}'\) are updated based on the features of the connected nodes and the edge itself:

\begin{equation}
   \ee_{ij}'' = \phi^{(l)}(\vv_i', \vv_j', \ee_{ij}') \quad \text{where} \quad \ee_{ij}'' \in \mathbb{R}^H.
\end{equation}

The node feature vectors \(\vv_i'\) are updated using the aggregated messages from their neighbors, \(\mathcal{N}(i)\), as follows:
\begin{equation}
   \vv_i'' = \gamma^{(l)}\left(\vv_i', \sum_{j \in \mathcal{N}(i)} \ee_{ij}''\right) \quad \text{where} \quad \vv_i'' \in \mathbb{R}^H.
\end{equation}

Unlike the Encoder module, the message-passing module explicitly uses the graph's structural information by incorporating the set of neighbors, \(\mathcal{N}(i)\), for each node. The number of message-passing steps, \(L\), is a hyperparameter that determines how far information propagates across the graph.

Residual connections are incorporated into the MLPs \(\phi\) and \(\gamma\). Additionally, the same parameters for \(\phi\) and \(\gamma\) are reused across all \(L\) message-passing steps. After completing \(L\) steps, the resulting graph \(\mathcal{M}'' = (\mathcal{V}'', \mathcal{E}'')\) is passed to the decoder.

\textbf{Decoder}: The decoder projects the latent node feature vectors into the desired output space. It uses an MLP, denoted $\delta^V$, to map each node feature vector $\vv_i''$ to an output vector $\bm{o}_i$ as follows:

\begin{equation}
    \delta^V(\vv_i'') \mapsto \bm{o}_i \quad \text{with} \quad \bm{o}_i \in \mathbb{R}^O.
\end{equation}

Here, $O$ represents the dimension of the output vector associated with each node. In this study, $O = 3$ since we aim to predict the local stress field components $(\hat{\sigma}_{xx_i}, \hat{\sigma}_{yy_i}, \hat{\sigma}_{xy_i})$ at each node of the input mesh $\mathcal{M}$.

In this implementation, we adopted the hyper-parameters from the original \textit{MeshGraphNets} framework as detailed in \cite{ref:mesh_graph_net}, specifically setting the latent space dimension $H$ to $128$ and the number of message-passing steps $L$ to $10$.

\subsection{Physics-Informed Loss} \label{sec:divergence_loss}

To guide the training process, we introduce a physics-informed loss function that enforces the equilibrium condition described in Section \ref{sec:divergence}. The total loss is composed of two terms: (1) a normalized mean squared error (\textrm{NMSE}) between the GNN predictions and the ground truth, and (2) a physics-based regularization term that penalizes deviations from the divergence-free stress condition $\mbox{div} \, \sigma = 0$.

Let $\sigma$ denote the ground-truth stress field obtained from FE simulations, $\hat{\sigma}$ the predicted stress field from the GNN, and $n$ the total number of nodes in the mesh $\mathcal{M}$. The loss function is defined as:

\begin{equation}
    \mathcal{L}\,(\sigma, \hat{\sigma}) = \textrm{NMSE}\, (\sigma, \hat{\sigma}) + \lambda \, (\mathbf{div}(\boldsymbol{\hat{\sigma}}))^2 .
\end{equation}

The physics-informed term  $(\mathbf{div}(\boldsymbol{\hat{\sigma}}))^2$ is computed using the divergence discrete operator (see Equation \eqref{eq:div_operator}, Section \ref{sec:divergence})
and is defined as the mean squared norm of the divergence over all indices $i$:

\begin{equation}
(\mathbf{div}(\boldsymbol{\hat{\sigma}}))^2 =
\frac{1}{n} \sum_{i=1}^n \| \left. \mathbf{div}(\boldsymbol{\hat{\sigma}}) \right|_{\mathbf{x}=\mathbf{x}_i}\|^2,
\label{eq:div_term}
\end{equation}

It is important to note that the divergence term in Equation \eqref{eq:div_term} is minimized using only internal nodes of the mesh (see blue nodes in Figure \ref{fig:graph_repr_schema}). This approach is necessary because the divergence operator applied to the stress field $\boldsymbol{\hat{\sigma}}$ is not well-suited for periodic meshes. In both, our model and FEM simulations, divergence values at the boundary nodes are significantly high. To address this issue, the divergence at these boundary nodes is set to zero during training.

The \textrm{NMSE} is used instead of the mean squared error (MSE) to ensure equal importance for each stress component, $c \in \{xx, yy, xy\}$. It is defined as:

\begin{gather}
    \textrm{NMSE}\, (\sigma, \hat{\sigma}) = \frac{1}{n_c} \sum_{c=1}^{n_c} \frac{\sum_{i=1}^{n} (\sigma_{c,i} - \hat{\sigma}_{c,i})^2}{\sum_{i=1}^{n} (\sigma_{c,i} - \frac{1}{n} \sum_{j=1}^n \sigma_{c,j})^2} ,
\end{gather}\label{eq:nmse_formulation}

The \textrm{NMSE} ensures that all stress components contribute equally to the training loss and the physics-based regularization term further encourages the GNN to satisfy the equilibrium condition.

A detailed analysis of the model's convergence and performance is presented in Sections \ref{sec:experimental_results_linear} and \ref{sec:experimental_results_non_linear}.

\section{Experimental results on linear elastic case} \label{sec:experimental_results_linear}

To validate the proposed approach and evaluate the contribution of the divergence-free regularization introduced in Section \ref{sec:divergence_loss}, we consider two datasets corresponding to different mechanical regimes. The first dataset consists of FEM simulations under purely linear elastic conditions. This configuration, being computationally straightforward, serves as a baseline for assessing the effectiveness of the physics-informed regularization during training.

To further investigate the efficiency and generalization capabilities of the model, particularly the benefits of employing a Graph Neural Network (GNN) as a localization operator, a second, more complex dataset is introduced. This dataset is derived from a non-linear hyperelastic FEM model that includes geometric nonlinearities. The corresponding experiments and results are presented in detail in Section \ref{sec:experimental_results_non_linear}.

\subsection{Dataset generation} \label{sec:dataset_linear_elastic}

\begin{figure}[ht]
    \centering
    \includegraphics[width=0.9\textwidth]{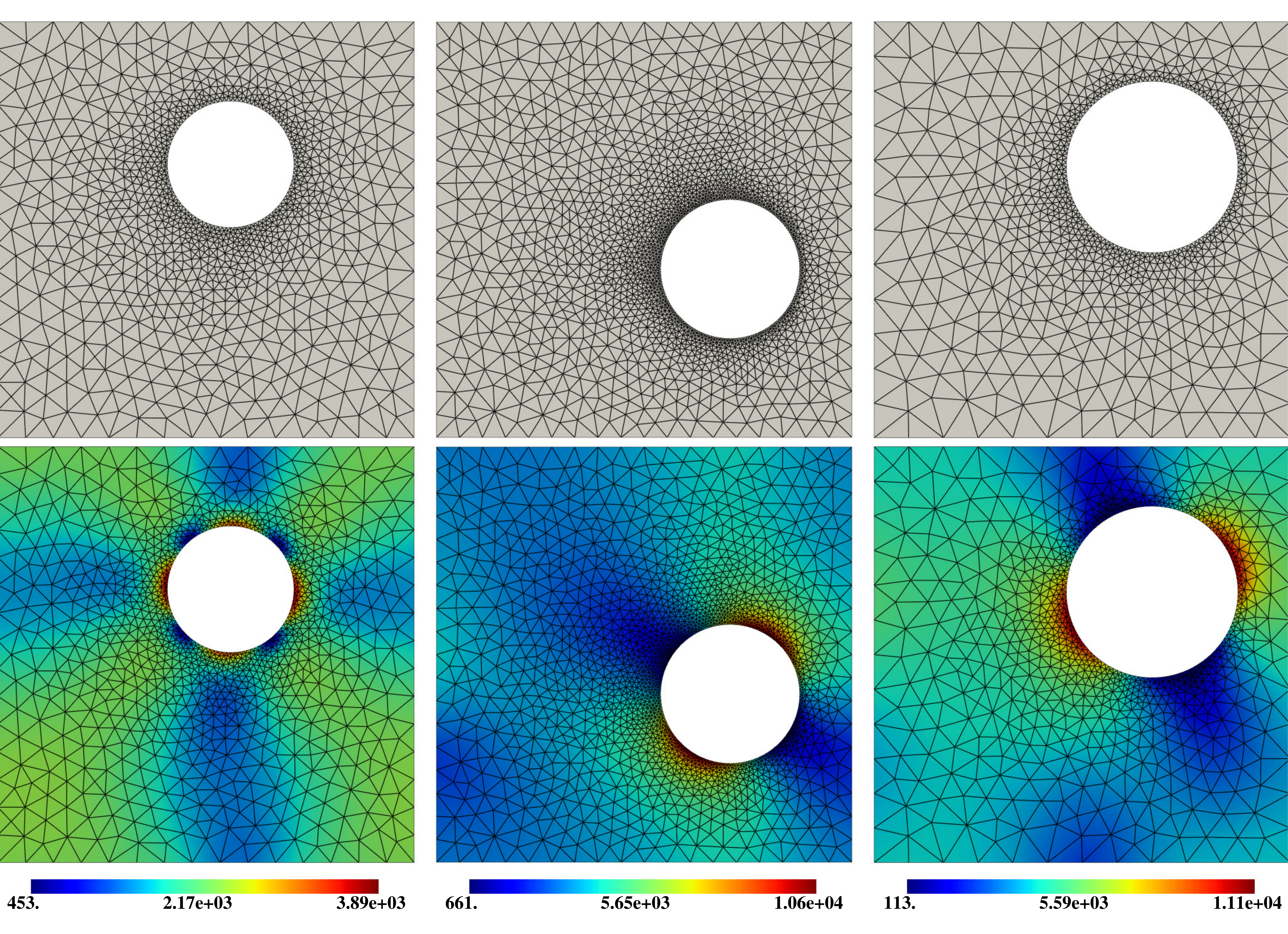}
    \caption{Top row: mesh examples of the generated database note that a mesh refinement is applied to every elements belonging to the hole surface, the mesh refinement and the global size of the triangles are different for each mesh. For this three examples the global triangles sizes corresponding to each mesh are $(7.10, 5.79, 8.35)$ and the triangle sizes belonging to the hole surface are $(1.01, 0.62, 1.07)$
      Bottom row: Von Mises local stress fields (in MPa) corresponding to top row meshes.}

    \label{fig:random_meshes}
\end{figure}

In this work, we generate a dataset consisting of 10,000 two-dimensional periodic meshes of a square
plate with a central hole. Mesh refinement is applied to every element
belonging to the central hole's surface. The meshes are generated using the
open-source meshing library \textit{Gmsh} \cite{ref:gmsh}. To introduce
variability, the coordinates of the hole's center, its radius, and the mesh
refinement levels are uniformly sampled (see Figure \ref{fig:random_meshes}),
resulting in meshes with different hole positions, radii, and triangle element
sizes. Each mesh is paired with a local stress field computed using Finite
Element simulation, as detailed in Section
\ref{sec:mechanical_problem}. Each solution is computed under periodic boundary
conditions applied to the edges of the square plate, assuming plane stress
conditions. The average in-plane strain components, \(\bar{\varepsilon}_{xx}\),
\(\bar{\varepsilon}_{yy}\), and \(\bar{\varepsilon}_{xy}\), are used to define
these periodic boundary conditions. To ensure the database captures a wide
range of loading conditions, mean in-plane strain values are uniformly sampled
within the range \([-0.05, 0.05]\).

The dataset, comprising 10,000 meshes with varying hole plate geometries and mesh refinements along with their corresponding FEM-computed solutions, is partitioned into a training set $(70\%)$ and a test set $(30\%)$. All results presented in this work are obtained exclusively from the test set.

Each simulation is performed using the
open-source \textrm{FE} Python library, \textit{Fedoo} \cite{ref:3mah}, with
local stress values computed at the Gauss points of the elements.
Considering the variability of the hole position and radius, note that
the formulated problem is non-linear even if linear elastic response is considered.
Indeed, the mapping function providing hole position, radius, average strain as
prescribed boundary conditions and linear elastic material parameters to get
the mechanical field over the domain is a non-linear function.

\FloatBarrier

The goal of the GNN is to reconstruct local stress
fields based on their average values. During training, the input stress values corresponds to
the volume average of stress fields over the representative unit
cell (RUC) volume, as expressed by:

\[
\bar{\sigma} = \frac{1}{V} \int_{\Omega} \sigma \, dV .
\]

The GNN is designed to predict stress values at the nodes of a graph, which
coincide with the structure of the FE mesh. The stress value shall be unique for each node,
consequently, the finite element stress values at nodes are computed as the average of interpolated
stress values from the surrounding elements.

A linear elastic isotropic material model was used, characterized by a Young’s
modulus of \(10^5\) MPa and a Poisson ratio of 0.3. A linear elastic constitutive
law was chosen to focus on evaluating the effects of the proposed
divergence-free regularization in terms of normalized mean squared error (\textrm{NMSE})
convergence, the divergence of predicted fields, and reconstruction quality considering a variation of the
geometric features of the problem, i.e. the graph structure.
Non-linear constitutive response can be considered as an extension of such proposed methodology, considering a sequence
of linearized increments of the macroscopic quantities.

To standardize the contributions of different stress components during
training, feature scaling was applied by subtracting the mean and dividing by
the standard deviation of the training dataset.

\subsection{Model training}\label{sec:training_elastic}

To benchmark the proposed models, \textit{\periodicModelName} and \textit{\proposedModelName}, are both trained for 200 epochs using the same training dataset and identical random seed initialization to ensure reproducibility. Model optimization is performed using the Adam algorithm \cite{ref:adam}, with a learning rate of $10^{-2}$. A mini-batching strategy is employed during training, using a batch size of 16 graphs per iteration. Early stopping is applied with a patience of 20 epochs based on the normalized mean squared error (NMSE) computed on the validation set. All experiments are conducted using the open-source library PyTorch Geometric \cite{ref:torch_geometric} and executed on an NVIDIA RTX A4000 GPU with 16 GB of VRAM.

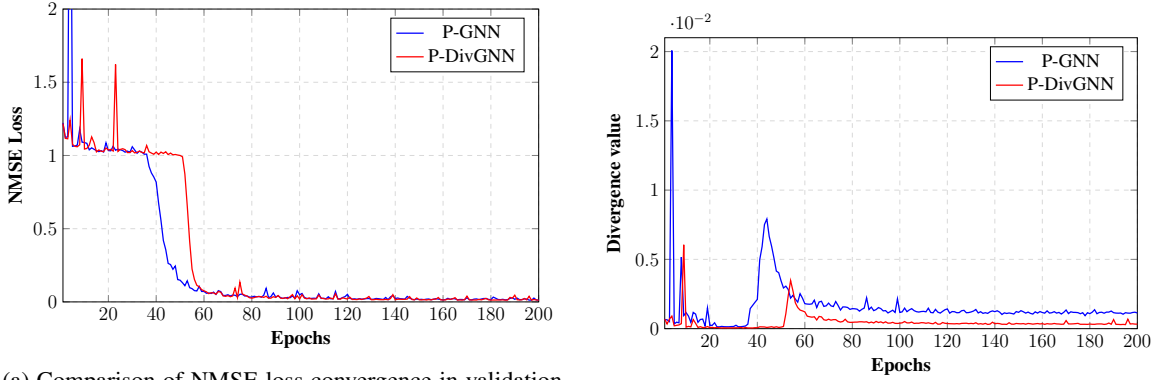
\begin{figure}[ht]
    \centering
    \begin{subfigure}{0.45\textwidth}
        \centering
        \resizebox{\linewidth}{!}{\begin{tikzpicture}
    \begin{axis}[
        width=12cm, height=8cm,
        grid=major,
        xlabel={Epochs},
        ylabel={NMSE Loss},
        legend pos=north east,
        label style={font=\bfseries\large},
        tick label style={font=\bfseries\large},
        legend style={nodes={scale=1.2, transform shape}},
        grid style={dashed, gray!30},
        xmin=1, xmax=200,
        ymin=0, ymax=2,
    ]

    \addplot[color=blue, thick]
        table [x index=0, y index=1, col sep=comma] {vanilla_gnn_nmse_val.csv};
    \addlegendentry{P-GNN}

    \addplot[color=red, thick]
        table [x index=0, y index=1, col sep=comma] {divgnn_nmse_val.csv};
    \addlegendentry{P-DivGNN}

    \end{axis}
\end{tikzpicture}}
        \caption{Comparison of \textrm{NMSE} loss convergence in validation dataset when the proposed
        divergence free penalty is applied during training}
        \label{fig:loss_curves}
    \end{subfigure}
    \hspace{1em}
    \begin{subfigure}{0.45\textwidth}
        \centering
        \resizebox{\linewidth}{!}{\begin{tikzpicture}
    \begin{axis}[
        width=12cm, height=8cm,
        grid=major,
        xlabel={Epochs},
        ylabel={Divergence value},
        label style={font=\bfseries\large},
        tick label style={font=\bfseries\large},
        legend pos=north east,
        legend style={nodes={scale=1.2, transform shape}},
        grid style={dashed, gray!30},
        xmin=1, xmax=200,
        ymin=0, ymax=0.021,
    ]

    \addplot[color=blue, thick]
        table [x index=0, y index=1, col sep=comma] {vanilla_gnn_divergence_value_val.csv};
    \addlegendentry{P-GNN}

    \addplot[color=red, thick]
        table [x index=0, y index=1, col sep=comma] {divgnn_divergence_value_val.csv};
    \addlegendentry{P-DivGNN}

    \end{axis}
\end{tikzpicture}}
        \caption{Divergence evolution in validation dataset during training}
        \label{fig:divergence_evolution}
    \end{subfigure}
    \caption{Comparison of \textrm{NMSE} loss and divergence evolution during training of \textit{\periodicModelName} model (in blue) and \textit{\proposedModelName} model (in red).}
    \label{fig:loss_side_by_side}
\end{figure}
\FloatBarrier

In Figure \ref{fig:loss_curves}, the evolution of the \textrm{NMSE} loss during training is presented. It is evident that the \textrm{NMSE} for \textit{\periodicModelName} decreases significantly around epoch $40$. In contrast, the \textrm{NMSE} for \textit{\proposedModelName} shows a noticeable reduction at epoch $60$. This delayed behavior can be attributed to the addition of a new term in the loss function, which increases the complexity of learning the ground truth FE solution.

Furthermore, Figure \ref{fig:divergence_evolution} examines the evolution of the divergence values of the predicted stress fields during training for both models. During the initial 40 epochs, the divergence values remain near zero. This behavior occurs because the same mean stress values are uniformly assigned to each node in the input graph. While this results in lower divergence values, the corresponding stress fields are unsatisfactory when compared to the FE solution. The regularization term hyperparameter is set to $\lambda = 10$ for balancing the contribution between data reconstruction performed by the $\textrm{NMSE}$ and the divergence constraint.

It is important to note that the divergence values of the predicted stress fields using \textit{\proposedModelName} remain consistently lower than those obtained with \textit{\periodicModelName} for the majority of the training process.

\subsection{Results}\label{sec:results_linear}
In our experimental setup, we assess the impact of incorporating a divergence-free penalty into the loss function during training. To this end, we compare the performance of \textit{\periodicModelName} and \textit{\proposedModelName} against finite element (FE) simulation results. The evaluation includes histogram distributions of each stress component $(\sigma_{xx}, \sigma_{yy}, \sigma_{xy})$, visual reconstructions of the stress fields, and a normalized squared error (\textrm{NSE}) metric to quantify deviations from the FE baseline.

The \textrm{NSE} is adapted from the normalized mean squared error (\textrm{NMSE}) introduced in Equation~\ref{eq:nmse_formulation}, but it is evaluated locally basis using nodal stress differences. It is defined as

\begin{gather}
    \textrm{NSE}\, (\sigma_{c,i}, \hat{\sigma}_{c,i}) =
    \frac{ (\sigma_{c,i} - \hat{\sigma}_{c,i})^2 }
         { \sum_{k=1}^{n} (\sigma_{c,k} - \frac{1}{n} \sum_{j=1}^n \sigma_{c,j})^2 } ,
\end{gather}\label{eq:nse_formulation}

This formulation enables a consistent quantitative comparison between the GNN predicted stress fields and the corresponding FE reference fields, using the same normalization scheme employed during network optimization. Consequently, the difference plots directly reflect the error measure relevant to the model’s training objective.

\begin{figure}[ht]
    \centering
    \includegraphics[width=0.8\textwidth]{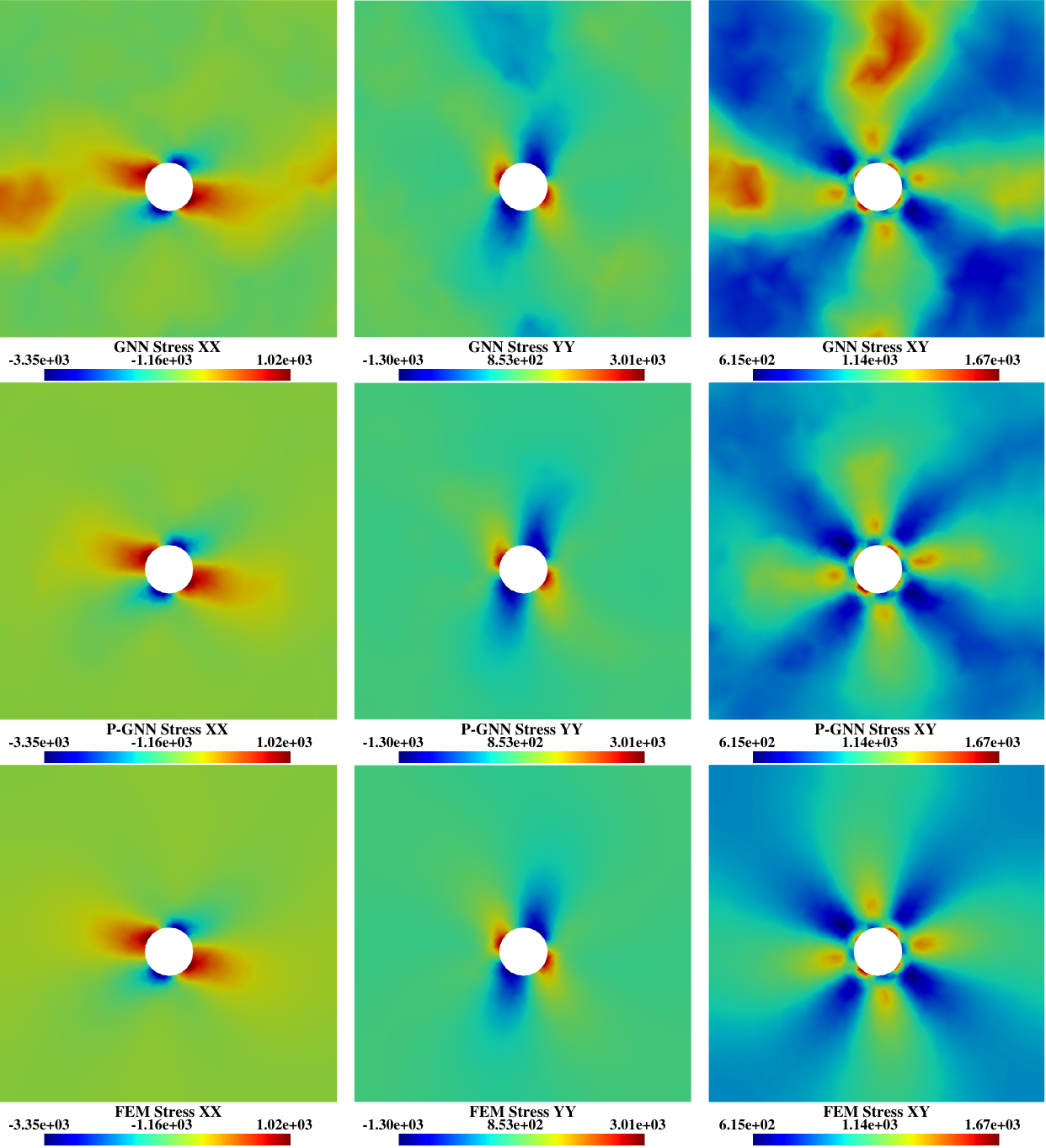}
    \caption{Local stress reconstruction comparison between classical GNN (without periodic edges) in top row, \textit{\periodicModelName} in middle row, and \textrm{FEM} in bottom row.}
    \label{fig:gnn_vs_p_gnn}
\end{figure} \FloatBarrier

\begin{figure}[ht]
    \centering
    \includegraphics[width=0.9\textwidth]{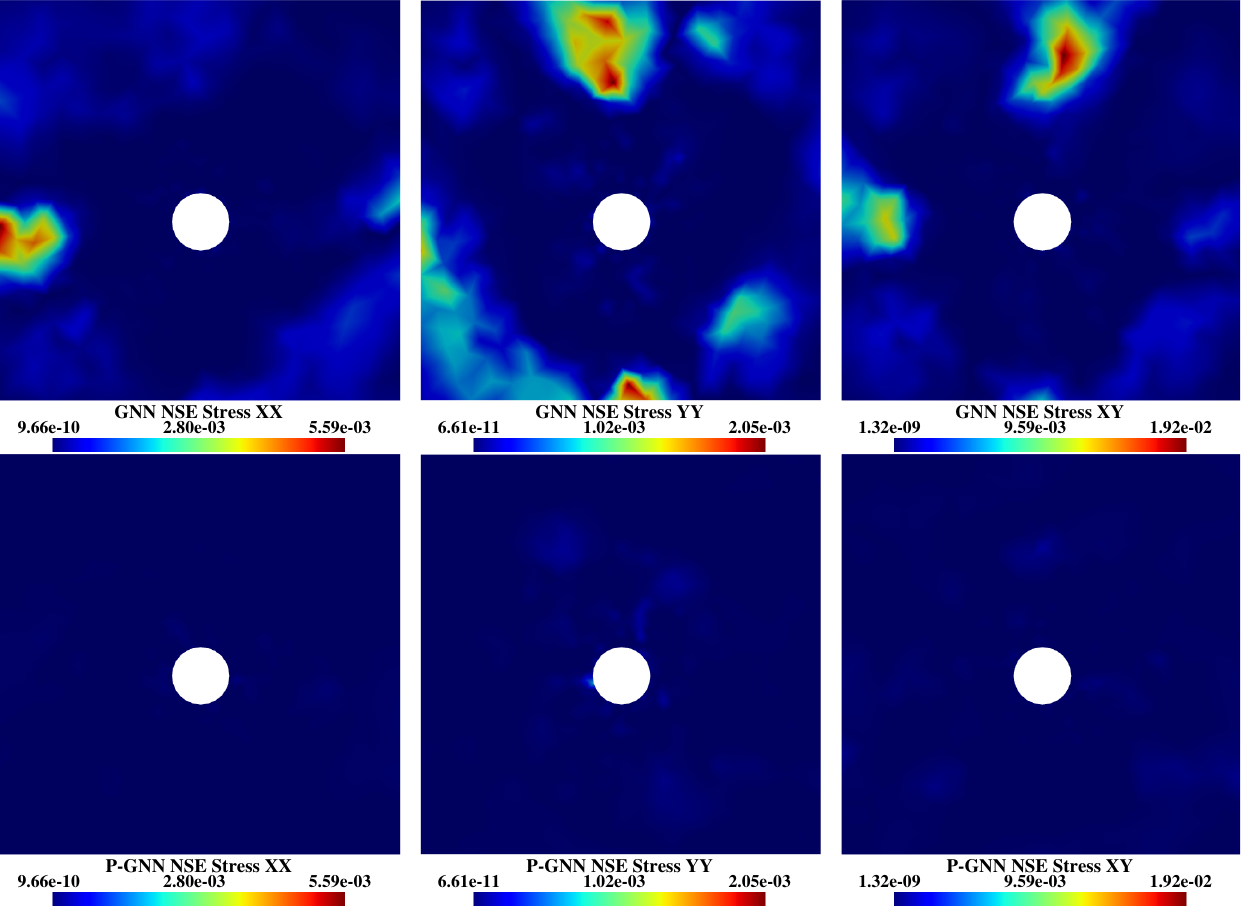}
    \caption{\textrm{NSE} between predicted local stress field and \textrm{FEM} stress field for both \textit{\periodicModelName} and classical GNN (without periodic edges), error fields are plotted at node level.}
    \label{fig:gnn_vs_p_gnn_nmse}
\end{figure}

As shown in Figure \ref{fig:gnn_vs_p_gnn} and Figure \ref{fig:gnn_vs_p_gnn_nmse}, the inclusion of periodic edges between boundary nodes leads to a marked improvement in the reconstruction of local stress fields. These additional periodic edges enhance the preservation of periodicity and symmetry in the predicted stress fields, resulting in more accurate reconstructions compared to a conventional GNN. Figure \ref{fig:gnn_vs_p_gnn_nmse} further emphasizes this effect through a difference plot that highlights the local reconstruction errors of the non-periodic GNN model.

This improvement is consistent across most of the test meshes. Consequently, the \textit{\periodicModelName} serves as the baseline model for further comparisons instead of the classical GNN framework.

 \FloatBarrier

It is important to note that this pre-processing step may introduce extra computational costs, particularly for very large meshes with a high number of nodes. If the representative volume element (RVE) has a squared shape (or cubic for 3D geometries), the operation is straightforward. In contrast, when the RVE does not have a regular shape, identifying boundary nodes may require the use of feature edge extraction algorithms, adding complexity to the process.

\FloatBarrier
\begin{figure*}[ht]
    \centering
     \hspace*{-2cm}
    \begin{subfigure}{\textwidth}
        \centering
         \includegraphics[width=1.0\textwidth]{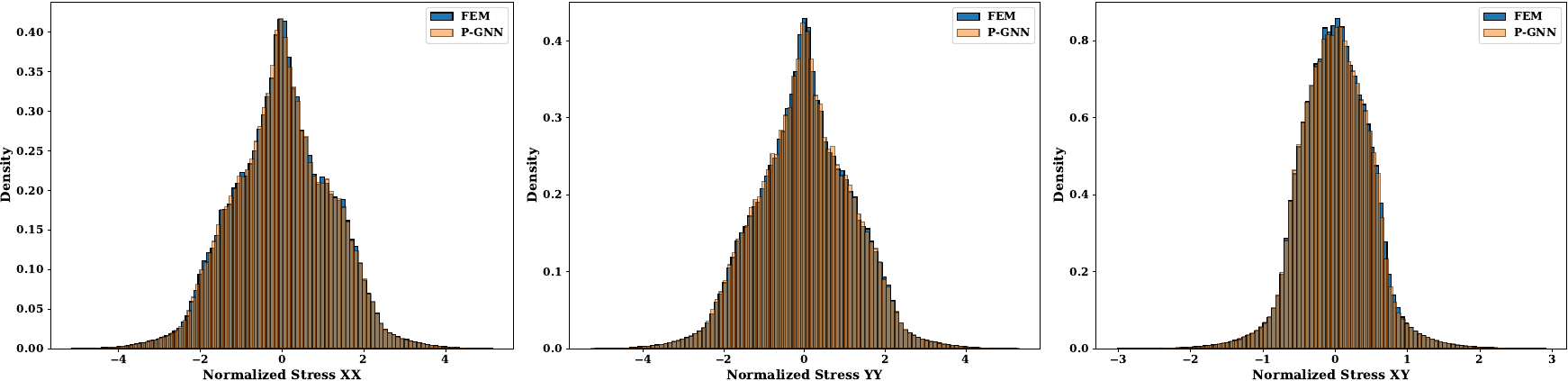}
        \caption{Comparison distribution of all local stress fields of the test database between predicted stress fields using \textit{\periodicModelName} and FE solutions}
        \label{fig:p_gnn_dist}
    \end{subfigure}
    \hspace{1em}
    \hspace*{-2cm}
    \begin{subfigure}{\textwidth}
        \centering
        \includegraphics[width=1.0\textwidth]{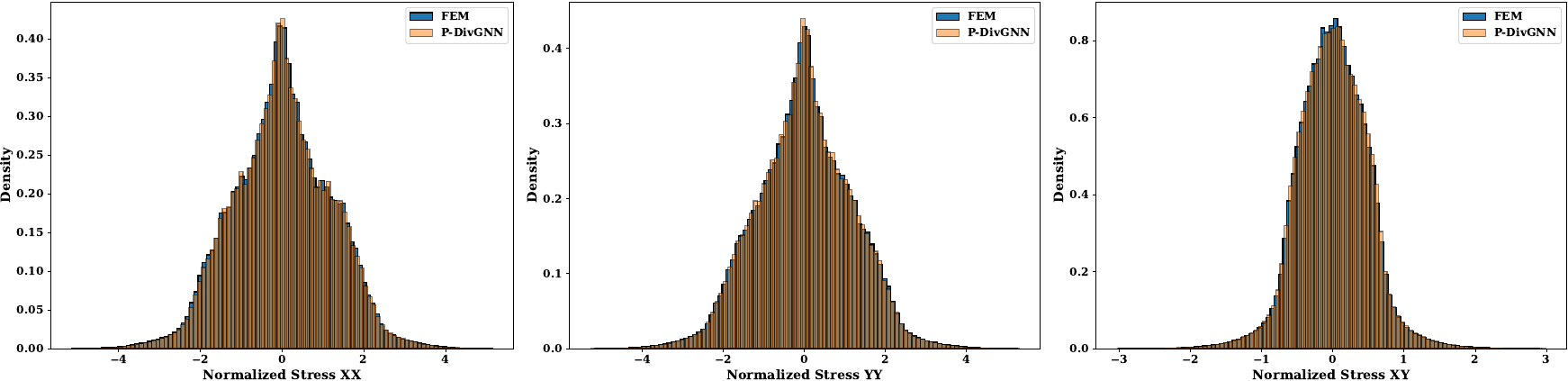}
        \caption{Comparison distribution of all local stress fields of the test database between predicted stress fields using \textit{\proposedModelName} and FE solutions}
        \label{fig:p_divgnn_dist}
    \end{subfigure}
    \caption{Comparison of normalized local stress distributions of the entire database between \textit{\periodicModelName}, \textit{\proposedModelName} and FEM.}
    \label{fig:distributions_compare}
\end{figure*}
\FloatBarrier

In Figure \ref{fig:distributions_compare}, we compare the standardized distributions of the predictions from both proposed models, \textit{\periodicModelName} and \textit{\proposedModelName}, against the \textrm{FEM} results. The stress fields for the entire test set, across each direction $(\sigma_{xx}, \sigma_{yy}, \sigma_{xy})$, are projected onto a density function. The histograms effectively illustrate that, overall, the predictions from both models are close to the FE ground truth. Especially, this visualization allows to determine if specific features of the stress distribution are captured, e.g., the repartition of extreme values of the stress fields. No truncation of the extreme stress values are observed in the predicted results.
Importantly, the incorporation of a divergence-free penalty does not degrade the stress values distribution but rather ensures that the predicted distributions remain consistent with the ground truth \textrm{FEM} solutions.

\begin{figure}[ht]
    \centering
    \includegraphics[width=0.9\textwidth]{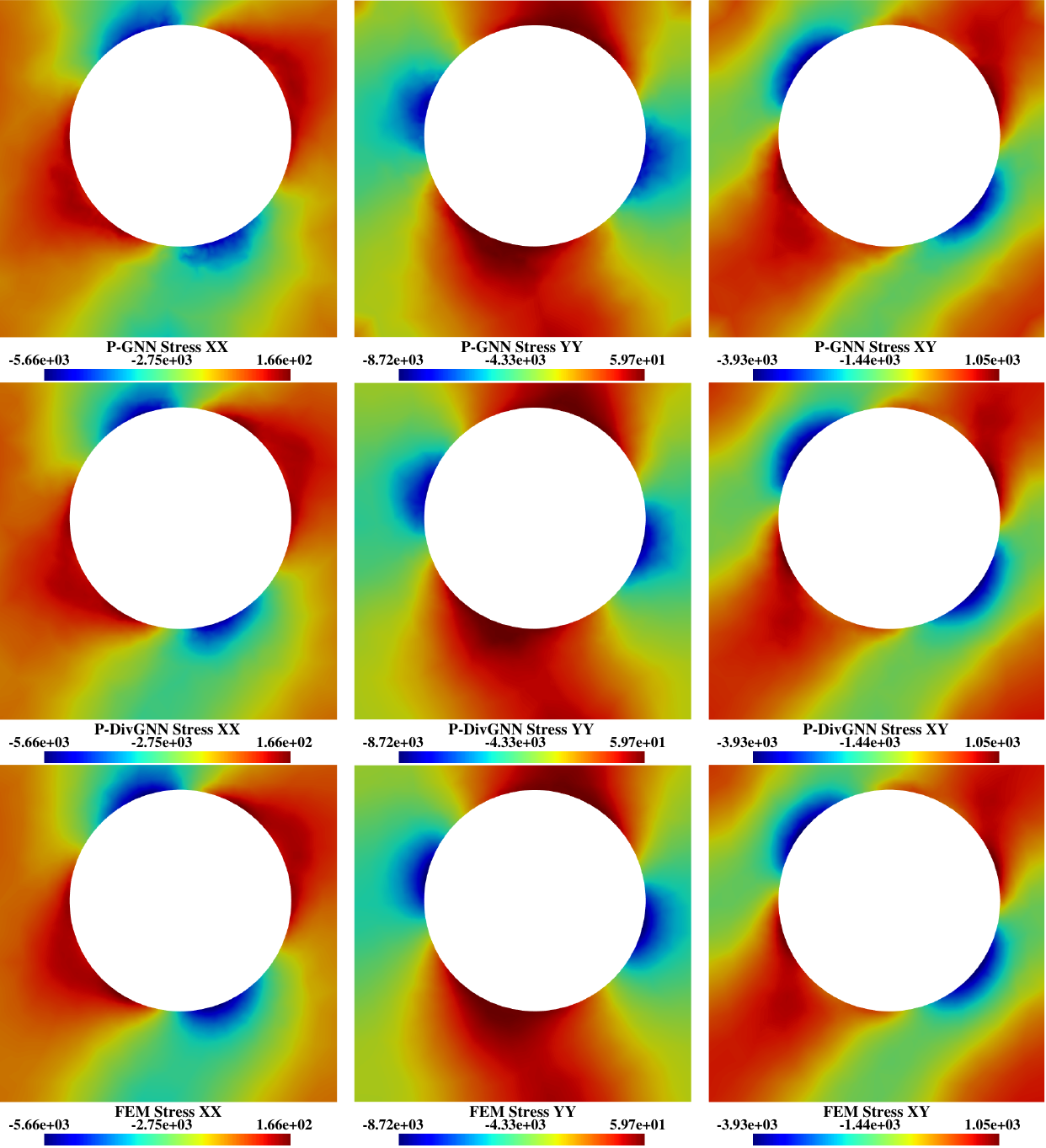}
    \caption{
    Comparison of local stress field reconstructions for \textit{\periodicModelName} (top row), \textit{\proposedModelName} (middle row), and \textrm{FEM} (bottom row) for the three stress components $(\sigma_{xx}, \sigma_{yy}, \sigma_{xy})$. Input mean stress values: $\bar\sigma_{xx} = -1154.90$ MPa, $\bar\sigma_{yy} = -2024.47$ MPa, $\bar\sigma_{xy} = -398.07$ MPa. The color bar scale is set to match the \textrm{FE} solution range.}
    \label{fig:compare_gnn_div_fem}
\end{figure} \FloatBarrier

Figure \ref{fig:compare_gnn_div_fem} compares the predictions of \textit{\periodicModelName}, \textit{\proposedModelName}, and the \textrm{FE} simulation for a representative example from the test set, which was not seen during training. In this example, \textit{\proposedModelName} produces noticeably smoother stress field predictions compared to \textit{\periodicModelName}, while both models generate reconstructions that are qualitatively comparable to the FE ground truth.

\begin{figure}[ht]
    \centering
    \includegraphics[width=0.9\textwidth]{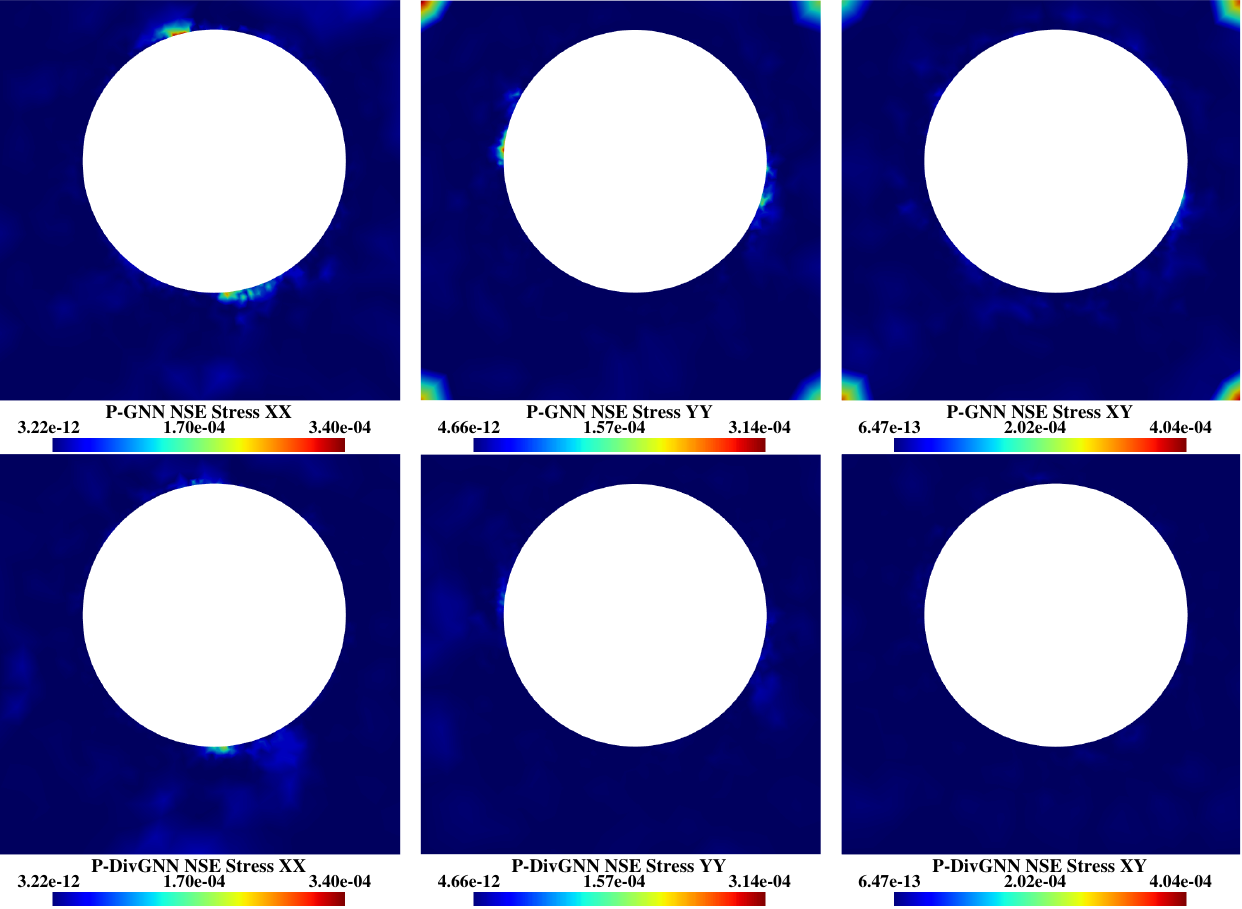}
    \caption{\textrm{NSE} between predicted local stress field and \textrm{FEM} stress field for both \textit{\periodicModelName}\  and \textit{\proposedModelName}, error fields are plotted at node level.}
    \label{fig:nmse_compare_div_no_div}
\end{figure}

For a more detailed comparison, Figure \ref{fig:nmse_compare_div_no_div} illustrates the \textrm{NSE} and error fields at the node level. Notably \textit{\periodicModelName} predicts higher stress values near the mesh corners, creating artifacts that are absent in the stress fields predicted by \textit{\proposedModelName}.

 \FloatBarrier

\begin{figure}[ht]
    \centering
    \includegraphics[width=1\textwidth]{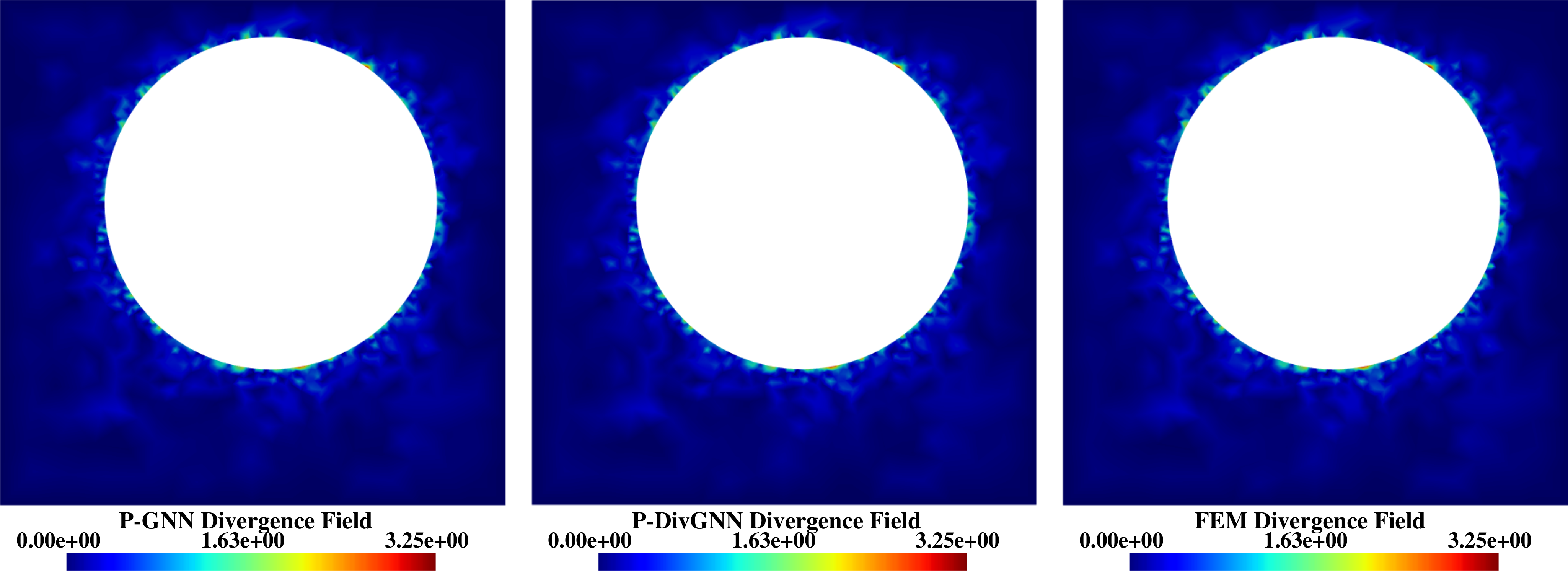}
    \caption{Comparison of norm of divergence field of local stress field between \textit{\periodicModelName}, \textit{\proposedModelName}\  and FEM predictions.}
    \label{fig:div_field_compare}
\end{figure} \FloatBarrier

Furthermore, the divergence field of the predicted stress fields is analyzed, as illustrated in Figure \ref{fig:div_field_compare}. The divergence field is computed as the norm of Equation \eqref{eq:div_operator} at each node of the mesh for the stress fields predicted by \textit{\periodicModelName}, \textit{\proposedModelName}, and the \textrm{FE} simulation.

As discussed in Section \ref{sec:divergence_loss}, Figure \ref{fig:div_field_compare} shows that the divergence values at the internal boundary nodes of the mesh are notably high.

It is important to note that the divergence operator used in this work does not account for periodicity when applied to external boundary nodes. Consequently, divergence values at these nodes (highlighted in red in Figure \ref{fig:graph_repr_schema}) are excluded from the training process and set to zero both during training and in the visualizations of Figure \ref{fig:div_field_compare}. The same exclusion applies to the nodes located along the hole boundary. This is because the stress values from the FE simulations are defined at the element level, and their divergence at the boundaries must be extrapolated using the derivatives of shape functions—an approximation deemed insufficiently accurate for training purposes.

Quantitatively, \textit{\proposedModelName} achieves a significantly lower divergence norm ($4.56 \times 10^{-4}$) than both the baseline models and the FEM reference solution, which yields a divergence of $9.43 \times 10^{-4}$. Moreover, the predicted stress field from \textit{\proposedModelName} also exhibits the lowest normalized mean squared error, reinforcing the effectiveness of the proposed approach in producing physically consistent stress predictions.

\begin{center}
\begin{table*}[!h]
\caption{Comparison of divergence mean value and \textrm{NMSE} between GNN (without periodic edges), \textit{\periodicModelName}, \textit{\proposedModelName}, and FEM, evaluated on the whole linear elastic test dataset. The model \textit{\proposedModelName} outperforms the baseline \textit{\periodicModelName} and achieves lower divergence values than the FE simulation.\label{tab:divergence_nmse}}

\begin{tabular*}{\textwidth}{@{\extracolsep\fill}lcc@{}}
\toprule
& \textbf{Divergence} & \textbf{NMSE} \\
\midrule
GNN        & $1.13 \times 10^{-3}$ & $2.33 \times 10^{-2}$ \\
P-GNN      & $1.11 \times 10^{-3}$ & $1.13 \times 10^{-2}$ \\
\textbf{P-DivGNN} & $\mathbf{3.23 \times 10^{-4}}$ & $\mathbf{1.08 \times 10^{-2}}$ \\
FEM        & $9.18 \times 10^{-4}$ & -- \\
\bottomrule
\end{tabular*}
\begin{tablenotes}
\item Results computed on the linear elastic test dataset.
\end{tablenotes}
\end{table*}
\end{center}
\FloatBarrier

Table \ref{tab:divergence_nmse} presents a quantitative comparison of the mean divergence values and NMSE for \textit{\periodicModelName}, \textit{\proposedModelName}, and FEM, evaluated over the entire test dataset. Specially, \textit{\proposedModelName} achieves a lower mean divergence value than FEM and a reduced NMSE compared to \textit{\periodicModelName}, demonstrating its improved accuracy and consistency despite adding an additional learning regularization.

This demonstrates the importance of incorporating the divergence regularization term during training, as it ensures a more physically plausible reconstruction of the stress fields.

\FloatBarrier
\subsubsection{Benchmark}

\begin{figure}[!htbp]
    \centering
    \includegraphics[width=0.9\textwidth]{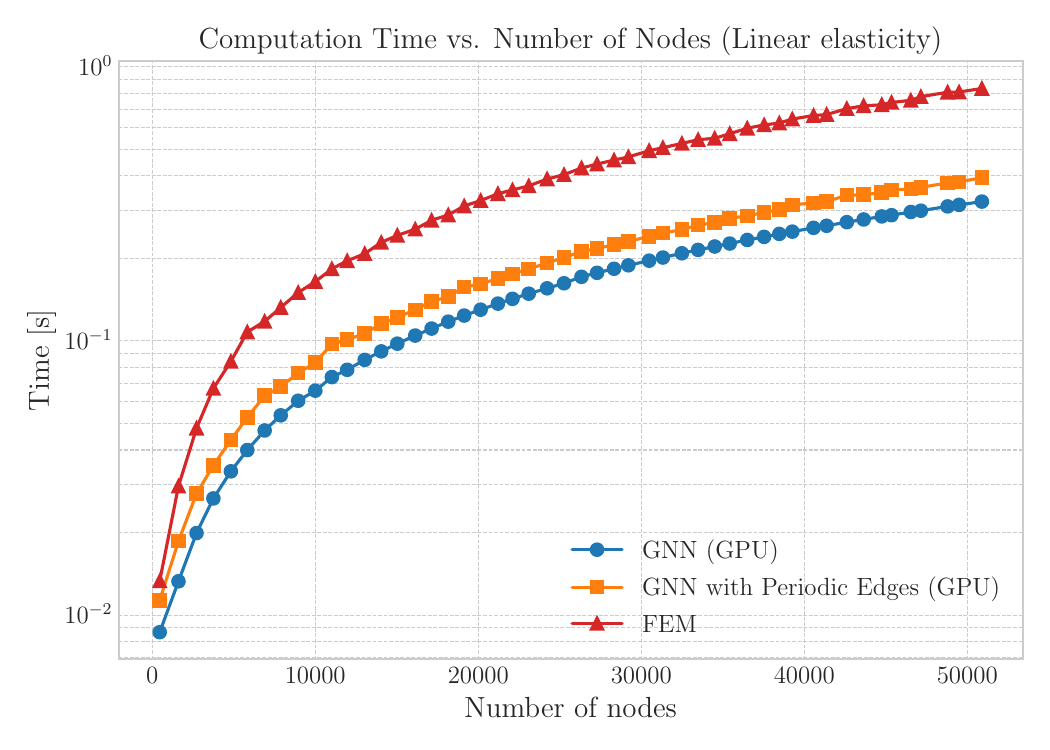}
    \caption{Performance comparison between \textrm{FEM} simulation \textit{\proposedModelName} (red), \textit{\proposedModelName} when adding periodic edges (orange) and without (blue). Tested using a CPU Intel Xeon W-$2255$ Processor, $10$ cores, $20$ threads for \textrm{FEM} simulation and GNN tests have been produced using a graphics card (GPU) Nvidia RTX A4000 with 16GB of VRAM where only 156MB were used.} \label{fig:benchmark}
\end{figure}

Figure \ref{fig:benchmark} presents a performance benchmark comparing the execution time of the proposed method (\textit{\proposedModelName}) in two configurations, one with periodic edges added and the other without, vs the execution time of a FE simulation. The benchmark is conducted by progressively increasing the number of nodes in the input mesh and measuring the computation time in seconds. Adding periodic edges introduces a computational cost due to the classification of nodes based on their association with a surface.

Although the inference time of \textit{\proposedModelName}\ is consistently lower than that of the FE simulation for all tested configurations, the comparison is not straightforward, as the GNN inference benefits from GPU acceleration (Nvidia RTX A4000, 16 GB VRAM), while the FE simulation runs on a CPU (Intel Xeon W-2255, 10 cores, 20 threads) using the Intel oneAPI Math Kernel Library PARDISO solver. Furthermore, the GNN requires a one-time training phase prior to inference, which is not reflected in the benchmark.

Still, these results are meaningful, even for a linear elastic problem where FE simulations are typically very fast, the proposed GNN achieves faster inference. This demonstrates the potential of the method, especially when moving toward more complex, nonlinear cases, as explored in Section {\ref{sec:benchmark_hyperelastic}} for a nonlinear hyperelastic case, where computational gains are expected to be even more significant.

\FloatBarrier
\section{Experimental results on non linear hyperelastic case}\label{sec:experimental_results_non_linear}

Building upon the results from the linear case (Section \ref{sec:experimental_results_linear}), we extend our evaluation to a dataset generated from a nonlinear hyperelastic FEM model that incorporates geometric nonlinearities. This more challenging scenario is designed to assess the robustness and performance gains achieved by using a GNN as a localization operator in complex mechanical settings.

As described in Section \ref{sec:mechanical_problem}, the FEM simulations for this case employ an iterative Newton-Raphson scheme, leading to a significantly higher computational cost compared to the linear case. In contrast, the proposed GNN model reconstructs the local stress field directly from the mean stress field at the RVE scale, without requiring iterative solvers. This enables substantial computational savings while maintaining accuracy in stress localization.

\subsection{Dataset generation} \label{sec:dataset_linear_hyperelastic}
The dataset generation procedure mirrors that of Section~\ref{sec:dataset_linear_elastic}. A total of $9795$ two dimensional periodic meshes of square plate with a hole featuring variability in hole geometry and mesh refinement, were generated using the same randomized sampling procedure. The dataset is partitioned into a training set $(70\%)$ and a test set $(30\%)$. As in the linear case, all results reported in this study are obtained from the test set.

Finite Element simulations were performed under periodic boundary conditions applied to the edges of the square plate. In contrast to the linear elastic case, which used plane stress and small strain assumptions, the hyperelastic simulations were conducted under plane strain and finite strain assumptions. This choice was motivated by improved convergence of the Newton–Raphson iterative solver during FEM computations. Importantly, this modeling assumption does not affect the generalizability or performance of the GNN, which remains agnostic to the underlying kinematic framework in terms of stress field reconstruction.

To cover a broader range of mechanical responses, the average in-plane logarithmic strain components \(\bar{\varepsilon}_{xx}\),
\(\bar{\varepsilon}_{yy}\), and \(\bar{\varepsilon}_{xy}\) were uniformly sampled within the range \([-0.15, 0.15]\).
 In this non linear model $\bar{\varepsilon}$ is the logarithmic strain tensor whereas it is the linearized small strain tensor in the linear one. Evidently, the two strain tensor measures are equivalent for small strain values.

As seen in Equation~\eqref{eq:finite_strain_bc}, for finite strain problems, the periodic boundary conditions are based on the displacement gradient $\overline{\nabla u}$.
The displacement gradients are computed from the sampled logarithmic strain tensors assuming a polar decomposition of $\mathbf{F}$ and enforcing an identity rotation matrix (i.e. without rigid body rotation).
This gives:
\begin{equation}
\overline{\nabla u} = \mathbf{F} -  \mathbf{1} = \exp{\bar{\varepsilon}} - \mathbf{1}
\end{equation}

The constitutive behavior is modeled using the Neo-Hookean hyperelastic law, as defined in Section~\ref{sec:mechanical_problem}, Equation~\eqref{eq:constitutive_law}, with the parameters: Shear modulus $\mu = 3.0$ MPa and bulk modulus $\kappa = 10 $ MPa.

\subsection{Model training}

To assess the convergence behavior of the models in the nonlinear hyperelastic setting, the same training protocol as described in Section \ref{sec:training_elastic} is adopted. Both \textit{\periodicModelName} and \textit{\proposedModelName} are trained for $200$ epochs using the Adam optimizer, with early stopping based on validation NMSE (patience of $20$ epochs), a learning rate of $10^{-2}$, and mini-batches of $16$ graphs. The primary difference in this setting is the choice of the physics-based regularization coefficient $\lambda$. Due to the different stress scales present in the hyperelastic dataset compared to the linear case, the regularization weight is adjusted to $\lambda = 10^{-2}$ to maintain an appropriate balance between the data fidelity and physics-informed terms in the loss function.

\begin{figure}[ht]
    \centering
    \begin{subfigure}{0.45\textwidth}
        \centering
        \resizebox{\linewidth}{!}{\begin{tikzpicture}
    \begin{axis}[
        width=12cm, height=8cm,
        grid=major,
        xlabel={Epochs},
        ylabel={NMSE Loss},
        legend pos=north east,
        legend style={nodes={scale=1.2, transform shape}},
        grid style={dashed, gray!30},
        label style={font=\bfseries\large},
        tick label style={font=\bfseries\large},
        xmin=1, xmax=200,
        ymin=0, ymax=2,
    ]

    \addplot[color=blue, thick]
        table [x index=0, y index=1, col sep=comma] {hyperelast_vanilla_gnn_nmse_val.csv};
    \addlegendentry{P-GNN}

    \addplot[color=red, thick]
        table [x index=0, y index=1, col sep=comma] {hyperelast_divgnn_nmse_val.csv};
    \addlegendentry{P-DivGNN}

    \end{axis}
\end{tikzpicture}}
        \caption{Comparison of \textrm{NMSE} loss convergence in validation dataset when the proposed
        divergence free penalty is applied during training}
        \label{fig:loss_curves_hyperelast}
    \end{subfigure}
    \hspace{1em}
    \begin{subfigure}{0.45\textwidth}
        \centering
        \resizebox{\linewidth}{!}{\begin{tikzpicture}
    \begin{axis}[
        width=12cm, height=8cm,
        grid=major,
        xlabel={Epochs},
        ylabel={Divergence value},
        label style={font=\bfseries\large},
        tick label style={font=\bfseries\large},
        legend pos=north east,
        legend style={nodes={scale=1.2, transform shape}},
        grid style={dashed, gray!30},
        xmin=1, xmax=200,
        ymin=0, ymax=0.08,
    ]

    \addplot[color=blue, thick]
        table [x index=0, y index=1, col sep=comma] {hyperelast_vanilla_gnn_divergence_value_val.csv};
    \addlegendentry{P-GNN}

    \addplot[color=red, thick]
        table [x index=0, y index=1, col sep=comma] {hyperelast_divgnn_divergence_value_val.csv};
    \addlegendentry{P-DivGNN}

    \end{axis}
\end{tikzpicture}}
        \caption{Divergence evolution in validation dataset during training}
        \label{fig:divergence_evolution_hyperelast}
    \end{subfigure}
    \caption{Comparison of \textrm{NMSE} loss and divergence evolution during training of \textit{\periodicModelName} model (in blue) and \textit{\proposedModelName} model (in red) in the hyperelastic dataset.}
    \label{fig:loss_side_by_side_hyperelast}
\end{figure}
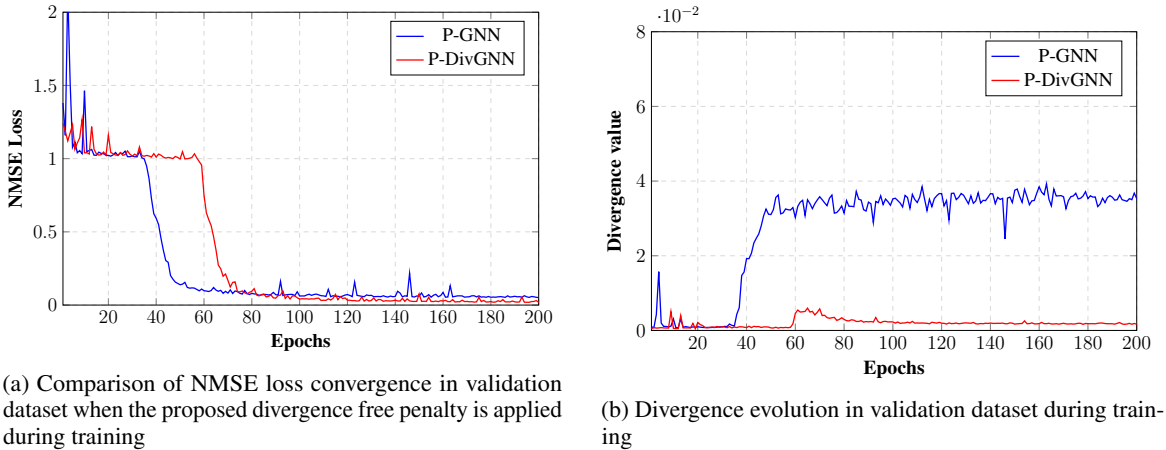 \FloatBarrier

The evolution of the NMSE loss during training on the nonlinear hyperelastic dataset is shown in Figure~\ref{fig:loss_curves_hyperelast}. A similar trend to that observed in the linear elastic case (Section~\ref{sec:training_elastic}) is noted: The \textit{\periodicModelName} exhibits a rapid decline in NMSE around epoch $40$, while \textit{\proposedModelName} demonstrates a delayed yet more progressive reduction, becoming significant near epoch $60$. Despite the slower initial convergence, \textit{\proposedModelName} ultimately achieves slightly lower NMSE values, suggesting that the inclusion of the physics-informed regularization term enhances the model's ability to reconstruct stress fields.

Figure~\ref{fig:divergence_evolution_hyperelast} illustrates the divergence evolution of the predicted stress fields throughout training. As in the linear case, divergence values are initially close to zero due to the uniform application of mean stress inputs across graph nodes. However, in this nonlinear scenario, the divergence profiles of the two models diverge more noticeably as training progresses. In particular, \textit{\proposedModelName} consistently maintains lower divergence values than \textit{\periodicModelName}, indicating improved compliance with the underlying physical constraints. This distinction is more pronounced than in the linear case and highlights the benefit of incorporating the divergence penalty, especially in more complex regimes where geometric nonlinearities are present.

\subsection{Results}

\begin{figure*}[ht]
    \centering
     \hspace*{-2cm}
    \begin{subfigure}{\textwidth}
        \centering
         \includegraphics[width=1.0\textwidth]{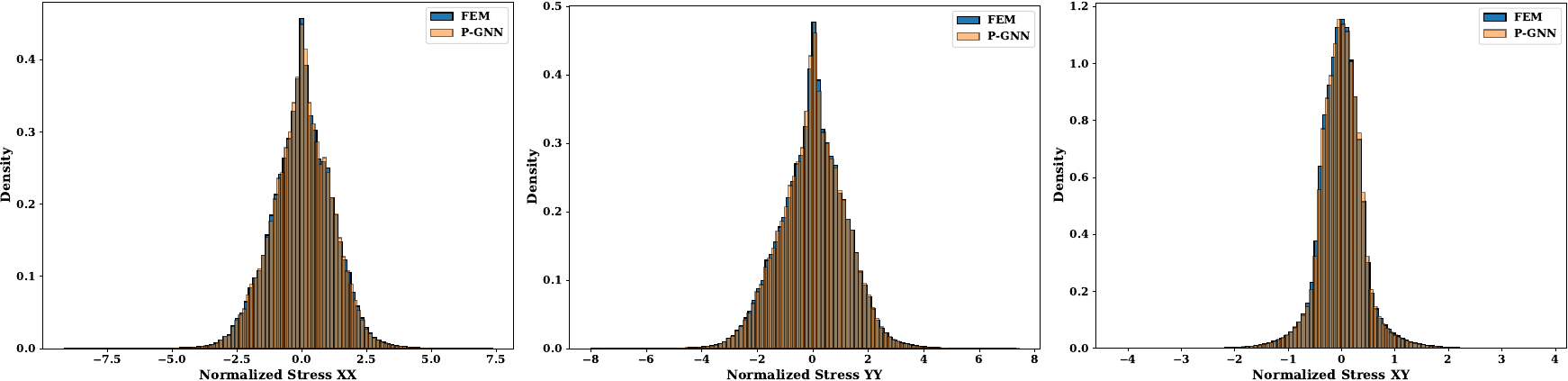}
        \caption{Comparison distribution of all local stress fields of the non-linear hyperelastic test database between predicted stress fields using \textit{\periodicModelName} and FE solutions}
        \label{fig:p_gnn_dist_hyperelast}
    \end{subfigure}
    \hspace{1em}
    \hspace*{-2cm}
    \begin{subfigure}{\textwidth}
        \centering
        \includegraphics[width=1.0\textwidth]{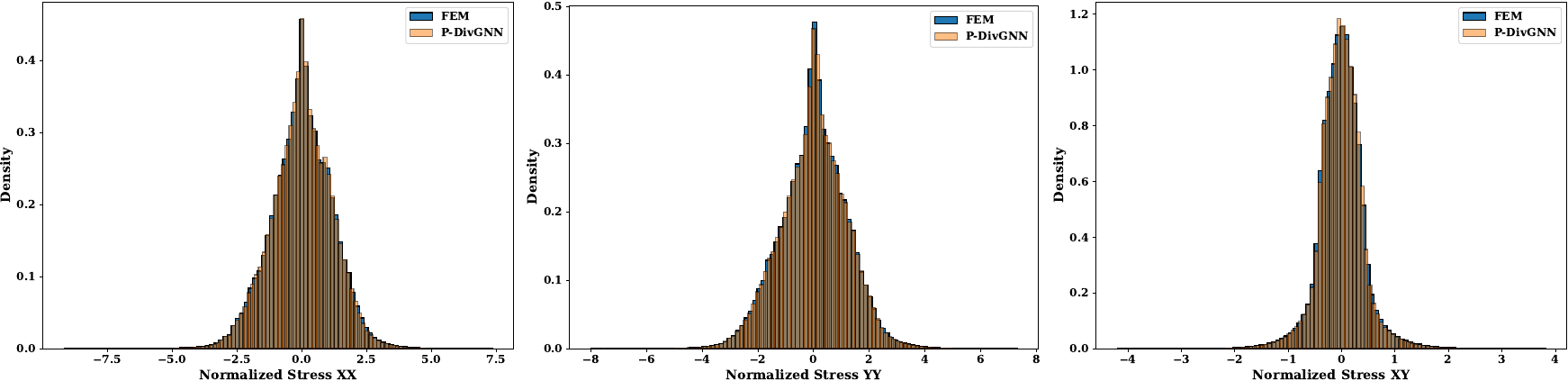}
        \caption{Comparison distribution of all local stress fields of the non-linear hyperelastic test database between predicted stress fields using \textit{\proposedModelName} and FE solutions}
        \label{fig:p_divgnn_dist_hyperelast}
    \end{subfigure}
    \caption{Comparison of normalized local stress distributions of the entire database between \textit{\periodicModelName}, \textit{\proposedModelName} and FEM.}
    \label{fig:distributions_compare_hyperelast}
\end{figure*}

Following the same evaluation strategy used for the linear elastic dataset, the performance of \textit{\periodicModelName} and \textit{\proposedModelName} is now assessed on the more complex nonlinear hyperelastic dataset. Figure~\ref{fig:distributions_compare_hyperelast} presents the distribution of predicted stress components $(\sigma_{xx}, \sigma_{yy}, \sigma_{xy})$ across the entire test set, compared to the reference FEM solutions.

\FloatBarrier

Figure \ref{fig:distributions_compare_hyperelast} shows that both models produce stress predictions that closely align with the ground truth distributions, preserving the overall statistical structure of the stress field. These results are particularly noteworthy given the increased complexity of the hyperelastic problem, as they indicate that the proposed GNN models are capable of accurately reconstructing stress fields even in the presence of geometric nonlinearities and non linear mechanical response.

The results indicate that the inclusion of the physics-informed regularization does not impair the fidelity of the predicted stress distribution. On the contrary, it contributes to maintaining consistency with the reference FEM results, even under nonlinear loading conditions. This demonstrates the robustness of the proposed approach in capturing realistic stress responses across a broader class of mechanical behaviors.

\begin{figure}[ht]
    \centering
    \includegraphics[width=0.9\textwidth]{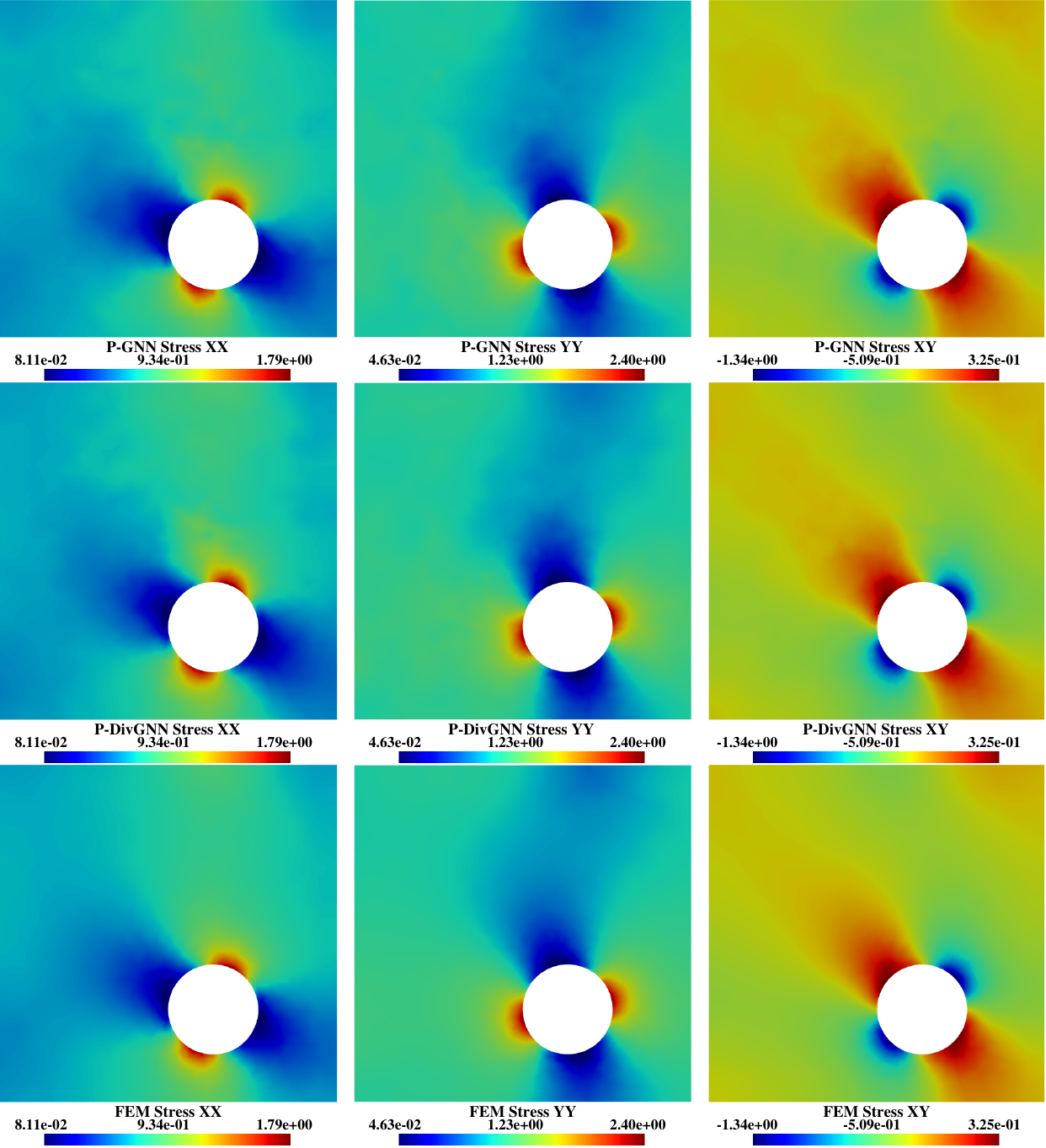}
    \caption{
    Comparison of local stress field reconstructions for \textit{\periodicModelName} (top row), \textit{\proposedModelName} (middle row), and \textrm{FEM} (bottom row) for the three stress components $(\sigma_{xx}, \sigma_{yy}, \sigma_{xy})$. Input mean stress values: $\bar\sigma_{xx} = 0.59$ MPa, $\bar\sigma_{yy} = -1.01$ MPa, $\bar\sigma_{xy} = 0.33$ MPa. The color bar scale is set to match the \textrm{FE} solution range.}
    \label{fig:compare_gnn_div_fem_hyperelast}
\end{figure} \FloatBarrier

Figure \ref{fig:compare_gnn_div_fem_hyperelast} presents a representative example from the hyperelastic test set. As observed in the linear elastic case (Section \ref{sec:results_linear}), both models provide stress field reconstructions that are qualitatively close to the FEM reference. However, the overall quality is slightly degraded in this non-linear setting. This is likely due to the increased complexity of the hyperelastic problem, which introduces geometric nonlinearities that are more challenging to learn. It is worth noting that the same GNN architecture and training settings were used for both datasets. Further performance improvements could be achieved through targeted hyperparameter tuning, such as optimizing the divergence penalty weight $\lambda$, but this was not explored in the current study.

\begin{figure}[ht]
    \centering
    \includegraphics[width=0.9\textwidth]{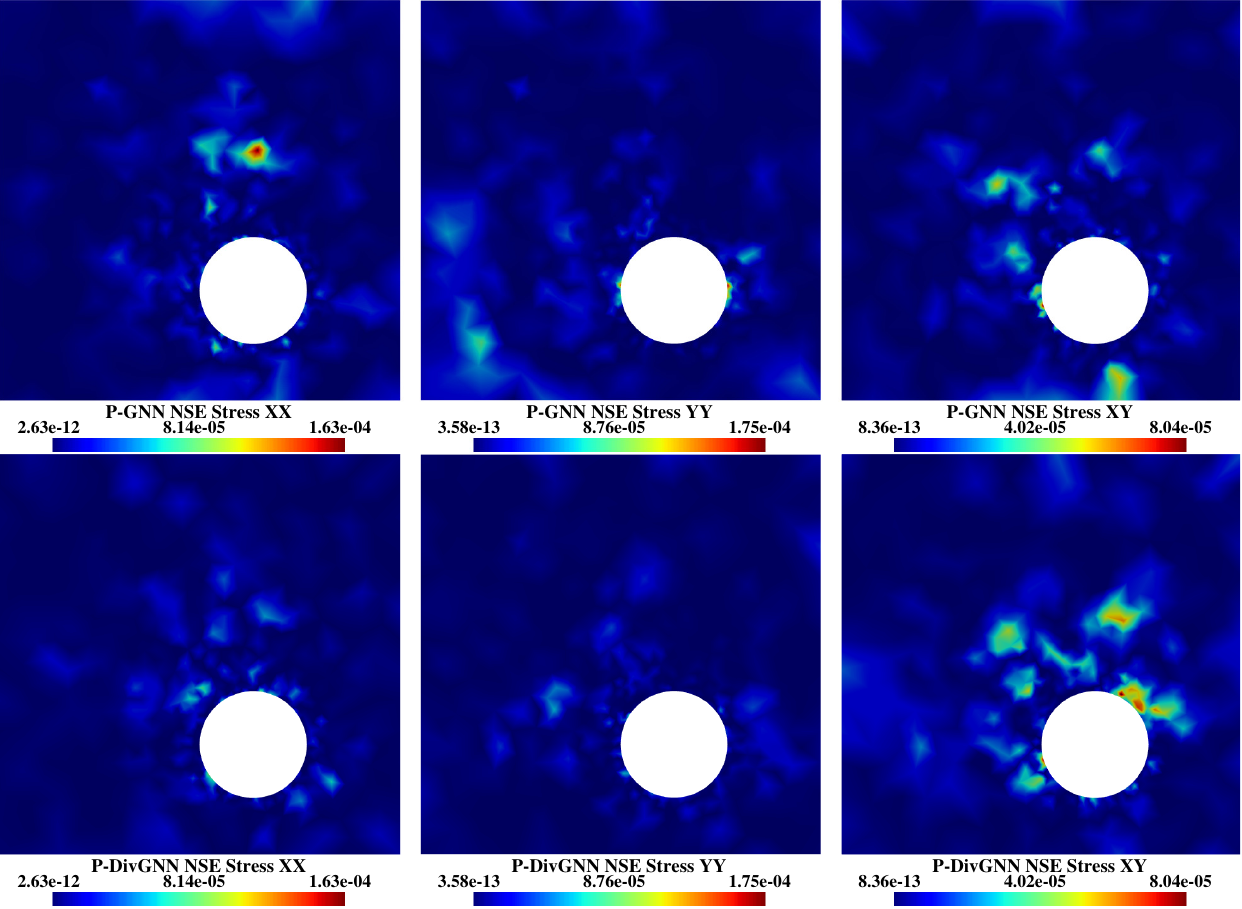}
    \caption{\textrm{NSE} between predicted local stress field and \textrm{FEM} stress field for both \textit{\periodicModelName}\  and \textit{\proposedModelName}, error fields are plotted at node level.}
    \label{fig:nmse_compare_div_no_div_hyperelastic}
\end{figure} \FloatBarrier

To complement the qualitative assessment, Figure \ref{fig:nmse_compare_div_no_div_hyperelastic} displays the corresponding node-level NSE error fields. The results indicate that \textit{\proposedModelName} yields lower local errors compared to \textit{\periodicModelName}, particularly in regions where the latter produces more irregular or noisy stress predictions. These observations confirm the benefit of incorporating a divergence constraint, especially in more complex, non-linear mechanical regimes.

\begin{figure}[ht]
    \centering
    \includegraphics[width=1\textwidth]{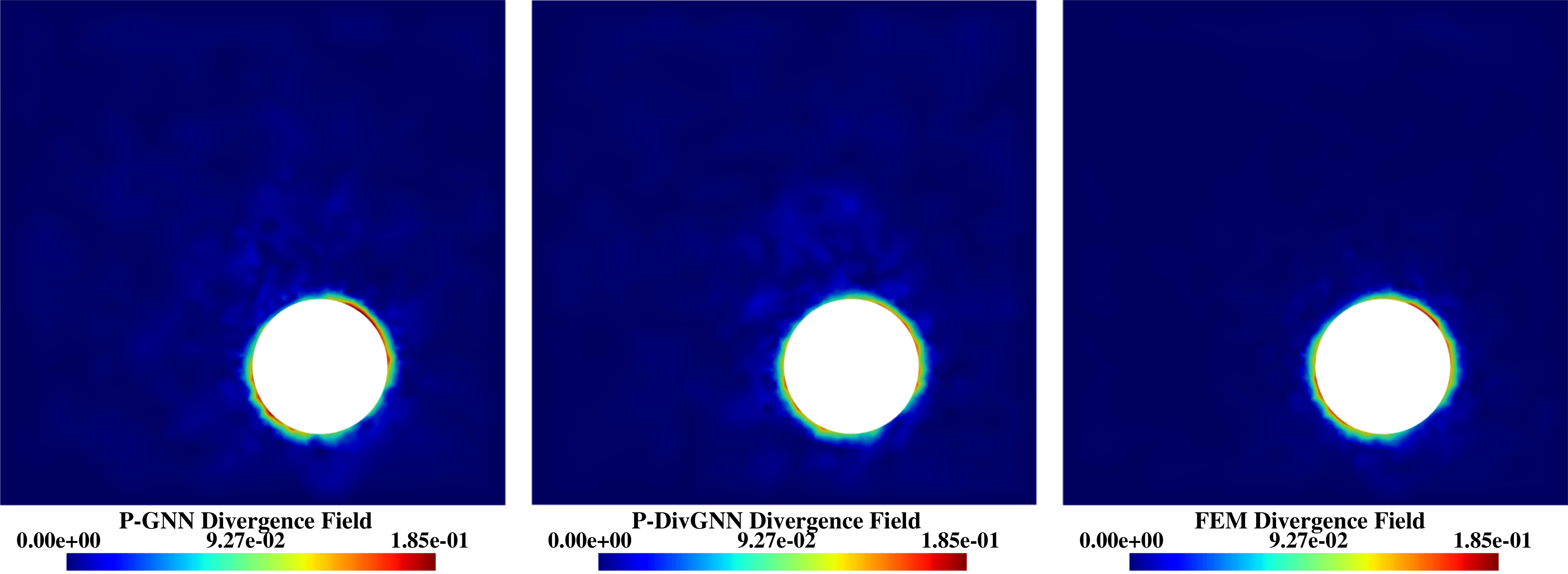}
    \caption{Comparison of norm of divergence field of local stress field between \textit{\periodicModelName}, \textit{\proposedModelName}\  and FEM predictions for a non-linear hyperelastic case.}
    \label{fig:div_field_compare_hyperelastic}
\end{figure} \FloatBarrier

Figure~\ref{fig:div_field_compare_hyperelastic} shows the divergence norm fields corresponding to the nonlinear hyperelastic stress predictions displayed in Figure~\ref{fig:compare_gnn_div_fem_hyperelast}. The divergence is computed for the stress fields predicted by \textit{\periodicModelName} and \textit{\proposedModelName}, as well as for the FEM reference solution. As in the linear case, the \textit{\proposedModelName} exhibits noticeably lower divergence values than the baseline model, particularly around the hole boundary where stress gradients are most pronounced. While FEM still achieves slightly lower divergence values in the nodes near the hole surface, the \textit{\proposedModelName} yields divergence values on the hole boundary itself that are even lower than those produced by FEM.

\begin{center}
\begin{table*}[ht]
\caption{Comparison of divergence mean value and \textrm{NMSE} between \textit{\periodicModelName}, \textit{\proposedModelName},
and FEM, evaluated on the whole non linear hyperelastic test dataset, the model \textit{\proposedModelName} achieves lower divergence values than \textit{\periodicModelName} and FE simulation.}
\label{tab:divergence_nmse_hyperelast}

\begin{tabular*}{\textwidth}{@{\extracolsep\fill}lcc@{}}
\toprule
& \textbf{Divergence} & \textbf{NMSE} \\
\midrule
        P-GNN      & $3.55 \times 10^{-2}$ & $5.36 \times 10^{-2}$ \\
        \textbf{P-DivGNN}   & $\mathbf{1.56 \times 10^{-3}}$ & $\mathbf{2.18 \times 10^{-2}}$ \\
        FEM        & $1.69 \times 10^{-3}$ & -- \\
\end{tabular*}
\begin{tablenotes}
\item Results computed on the non linear hyperelastic test dataset.
\end{tablenotes}
\end{table*}
\end{center}
\FloatBarrier

Table~\ref{tab:divergence_nmse_hyperelast} presents a quantitative comparison of the mean divergence and normalized mean squared error (NMSE) over the entire test set for the nonlinear hyperelastic dataset. The results confirm the advantage of incorporating a divergence-free penalty during training. Specifically, the proposed model \textit{\proposedModelName} achieves the lowest divergence value, outperforming both the baseline \textit{P-GNN} and even the FEM solution. This is a notable result, as it indicates that the model not only approximates the ground truth but also enforces physical consistency more effectively than the numerical solver under the same conditions. Additionally, the \textit{\proposedModelName} model yields significantly lower NMSE compared to \textit{\periodicModelName}, demonstrating improved fidelity in reconstructing the local stress fields. These metrics, taken together with the visual and error-field comparisons discussed earlier, provide robust evidence of the benefits introduced by the divergence-aware loss formulation.
\subsubsection{Benchmark}\label{sec:benchmark_hyperelastic}

\begin{figure}[!htbp]
    \centering
    \includegraphics[width=0.9\textwidth]{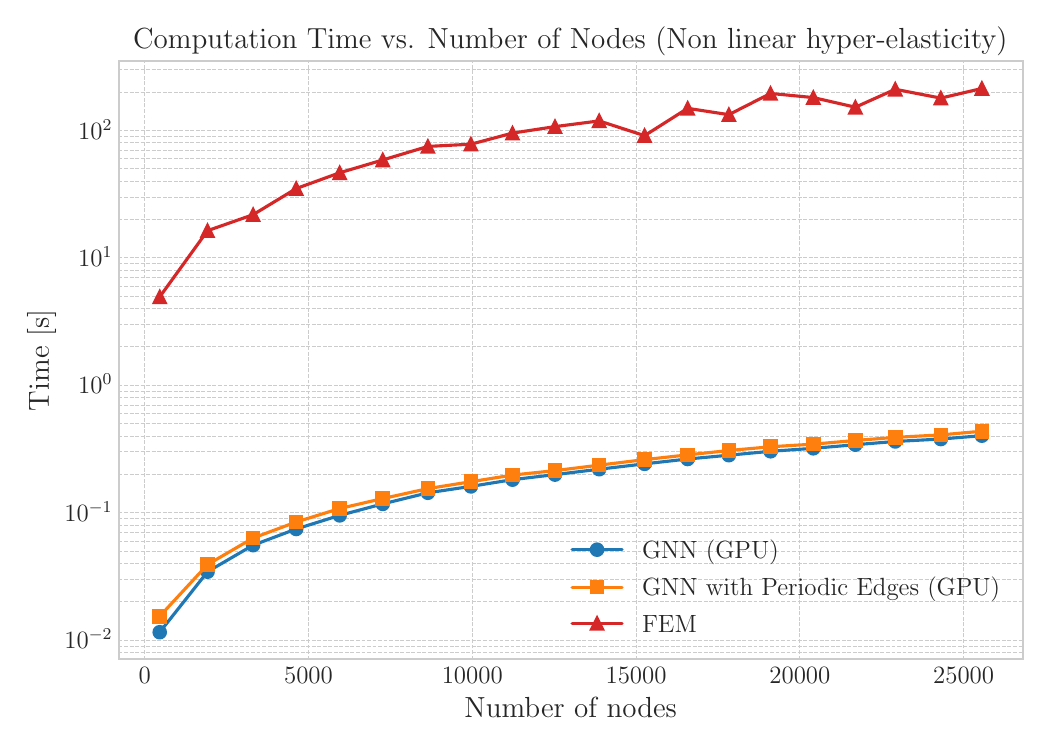}
    \caption{Performance comparison between hyperelastic \textrm{FEM} simulation \textit{\proposedModelName} (red), \textit{\proposedModelName} when adding periodic edges (orange) and without (blue). Tested using a CPU Intel Xeon W-$2255$ Processor, $10$ cores, $20$ threads for \textrm{FEM} simulation and GNN tests have been produced using a graphics card (GPU) Nvidia RTX A4000 with 16GB of VRAM where only 156MB were used.} \label{fig:benchmark_hyperelastic}
\end{figure}

Figure \ref{fig:benchmark_hyperelastic} presents the same benchmark test as in Figure \ref{fig:benchmark}, but applied to the reconstruction of a local stress field in a nonlinear hyperelastic problem. In this case, the FEM solution requires an iterative Newton–Raphson scheme which introduces variability in computation time regarding the mesh geometry and input strain tensors, the fluctuations on the FEM computation reported times are explained arise from adaptive time-stepping: When the convergence criteria are not met, the solver automatically reduces the time step, increasing the number of iterations required for getting the FEM solution.

The proposed model achieves a significant computational speed-up of approximately $\approx500$x primarily due to the absence of using an iterative scheme and the use of highly parallelized GPU matrix operations. The computation time of GNN scales similarly to the computation time of FEM in function of the number of nodes. It is important to note that the batch prediction capability of neural networks, allowing multiple problems to be solved simultaneously, was not utilized in this study. Therefore, the reported speed-up represents a conservative estimate. In a full FE² setting, where multiple RVEs are evaluated concurrently, even greater computational gains can be expected.

\FloatBarrier
\section{Conclusion and perspectives}

In this study, we developed a novel framework for reconstructing local stress fields from homogenized mean stress values obtained through ROM simulations. The proposed approach, named \textit{\proposedModelName}, leverages GNNs trained using a physics-informed loss function. This loss enforces equilibrium constraints during the learning process, ensuring that the predicted local stress fields minimizes the equilibrium condition $\mbox{div} \, \sigma = 0$. The framework enables the localization of stress fields within a representative unit cell (RUC) of a periodic microstructure. To achieve this, a graph is constructed from the input periodic mesh, incorporating periodicity by adding edges that connect opposing nodes of the mesh. Our results demonstrate that the inclusion of these periodic edges significantly enhances the accuracy of the predicted stress fields compared to a standard mesh to graph representation.

To evaluate the accuracy and robustness of \textit{\proposedModelName}, we conducted tests on two distinct mechanical scenarios, both involving two-dimensional plate geometries with a central hole. In each case, the hole’s position, radius, and mesh refinement were independently sampled from uniform distributions to introduce geometric variability. The first test case focused on linear elastic material behavior and was used to validate the model's predictive performance. The second test case involved a nonlinear hyperelastic material model and served to assess the model’s ability to reconstruct complex stress fields under large deformations. Additionally, this nonlinear scenario highlights the significant computational speed-up achieved by the proposed method compared to conventional finite element simulations.

The results indicate that \textit{\proposedModelName}\  generalizes well to unseen hole plate mesh configurations, successfully handling variations in node positions and mesh refinement. Comparative analysis with FE simulations revealed that \textit{\proposedModelName}\  is competitive in terms of both stress field reconstruction accuracy and computational efficiency. The impact of the physics-informed loss was further analyzed by comparing the divergence of the predicted stress fields with that of the FE-generated fields. On average, \textit{\proposedModelName}\  produces stress fields with lower divergence, indicating that, in some cases from the test database, the equilibrium state is better maintained using \textit{\proposedModelName}\  than with conventional FE simulations.

In future work, it would be valuable to explore the benefits of imposing a divergence-free constraint when localizing stress fields in the context of plasticity. Additionally, extending the proposed framework to handle more complex geometries and 3D cases could further expand its applicability. For temporal sequences of linearized increments, this approach could be adapted with minimal modifications, offering the potential for significant gains in computational efficiency.

\section{Code availability}

The source code allowing to generate the datasets used and to reproduce the experiments presented in this article can be found at: \\
\href{https://github.com/ricardo0115/p-div-gnn}{https://github.com/ricardo0115/p-div-gnn}

\section*{Author contributions}

\textbf{Manuel Ricardo Guevara Garban}: Writing – original draft,
Visualization, Validation, Software, Resources, Methodology, Investigation,
Formal analysis, Conceptualization. \textbf{Étienne Prulière}: Writing – review
and editing, Supervision, Software, Resources, Methodology, Conceptualization.
\textbf{Michaël Clément}: Writing – review and editing, Supervision, Software,
Resources, Methodology, Conceptualization. \textbf{Yves Chemisky}: Writing –
review and editing, Validation, Supervision, Software, Resources, Project
administration, Methodology, Investigation, Funding acquisition, Formal
analysis, Conceptualization.

\section*{Financial disclosure}

None reported.

\section*{Conflict of interest}

The authors declare no potential conflict of interests.

\bibliographystyle{unsrt}
\bibliography{biblio}

\end{document}